\documentclass[a4paper,fleqn]{cas-sc}

\usepackage{amsmath,amssymb,amsfonts,mathtools}
\usepackage{booktabs}
\usepackage{graphicx}
\usepackage{float}
\usepackage{subcaption}
\usepackage[table,xcdraw]{xcolor}
\usepackage{url}
\usepackage{array}
\usepackage{newfloat}
\usepackage[numbers]{natbib}
\usepackage[T1]{fontenc}

\DeclareFloatingEnvironment[name={Supplementary Figure}]{suppfigure}
\DeclareFloatingEnvironment[name={Supplementary Table}]{supptable}

\begin{document}
\let\WriteBookmarks\relax

\shorttitle{Morphology and actuation as inductive biases}
\shortauthors{Z. Tari et~al.}

\title[mode=title]{Morphology and actuation as inductive biases in robotic hand manipulation}

\author[itk]{Zal\'an Tari}[orcid=0009-0004-9133-7780]
\cormark[1]
\ead{tari.zalan@itk.ppke.hu}

\author[itk]{Eszter Birtalan}[orcid=0009-0009-9189-2481]

\author[itk]{P\'eter Polcz}[orcid=0000-0002-4217-0935]

\author[itk]{Mikl\'os Koller}[orcid=0000-0001-6283-114X]

\affiliation[itk]{organization={P\'azm\'any P\'eter Catholic University, Faculty of Information Technology and Bionics},
            city={Budapest},
            country={Hungary}}

\cortext[1]{Corresponding author.}

\begin{abstract}
Robotic hands vary widely in anatomical fidelity and mechanical complexity, and these structural choices influence the coordination of joint motions and the difficulty of controlling the system. A unified framework is presented in which the kinematic and actuation stages are analysed separately and in composition, through the conditioning of the task Jacobian, the actuation matrix, and their product. It is applied to two hands representing opposing design philosophies, the Shadow Dexterous Hand and the Anatomically Correct, Biomechatronic Hand, along four morphological aspects: joint axis geometry, actuator-to-DOF ratio, coupling architecture, and authority distribution. All parameters are derived from the hands' canonical digital representations. Anatomical fidelity carries no uniform advantage: oblique axes improve thumb conditioning but leave the long fingers worse conditioned than the orthogonal-axis design, while the branching tendon network improves the effective control mapping at every long finger and worsens it significantly at the thumb, where actuator authority is concentrated on thumb opposition. Predictions derived from these metrics are evaluated against reinforcement learning experiments using PPO, DDPG+HER, and TQC+HER, across three different tasks.
\end{abstract}



\maketitle

\section*{Nomenclature}
\noindent\begin{tabular}{@{}p{2.4cm}p{0.72\linewidth}@{}}
\multicolumn{2}{@{}l}{\textit{Symbols}}\\
$q$            & Joint configuration vector, $q\in\mathbb{R}^{n}$\\
$n$            & Number of kinematic degrees of freedom (DOF)\\
$m$            & Number of independent actuators\\
$k$            & Task-space dimensionality\\
$r$            & Kinematic redundancy, $r=n-k$\\
$x$            & Task-space vector (e.g.\ fingertip position/orientation)\\
$T(q)$         & Forward-kinematics transform, $T(q)\in SE(3)$\\
$T_0$          & Zero-configuration (home) pose\\
$\hat{\xi}$    & Joint screw axis, $\hat{\xi}=(\omega,v)\in\mathbb{R}^{6}$\\
$J(q)$         & Task Jacobian, $J\in\mathbb{R}^{k\times n}$\\
$A(q)$         & Actuation matrix, $\dot{q}=A(q)\,u$\\
$B(q)$         & Effective control mapping, $B(q)=J(q)A(q)$\\
$R_m(\theta)$  & Moment-arm Jacobian of the tendon network\\
$L_m$          & Vector of muscle-level tendon lengths\\
$u$            & Actuator input vector, $u\in\mathbb{R}^{m}$\\
$\sigma$       & Singular value of a matrix\\
$\kappa(\cdot)$& Condition number, $\sigma_{\max}/\sigma_{\min}$\\
$\omega(q)$    & Manipulability measure\\
$q_{\text{ref}}$& Mid-range reference configuration\\[4pt]
\multicolumn{2}{@{}l}{\textit{Abbreviations}}\\
SDH            & Shadow Dexterous Hand\\
ACBH           & Anatomically Correct, Biomechatronic Hand\\
DOF            & Degree(s) of freedom\\
PoE            & Product of Exponentials\\
D-H            & Denavit--Hartenberg\\
URDF           & Unified Robot Description Format\\
CMC / TCMC     & Carpometacarpal / trapeziometacarpal joint\\
MCP, PIP, DIP, IP & Metacarpophalangeal, proximal/distal interphalangeal, interphalangeal joints\\
RL             & Reinforcement learning\\
PPO            & Proximal Policy Optimization\\
DDPG           & Deep Deterministic Policy Gradient\\
TQC            & Truncated Quantile Critics\\
HER            & Hindsight Experience Replay\\
\end{tabular}

\clearpage

\section{Introduction}

\begin{figure}[H]
	\centering
	\begin{subfigure}[b]{0.48\linewidth}
		\centering
		\includegraphics[width=0.5\linewidth]{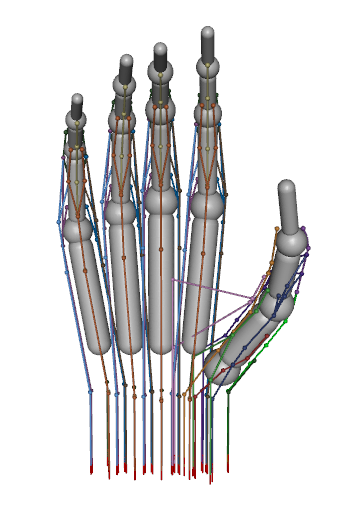}
		\caption{Simulated model of the Anatomically Correct, Biomechatronic Hand (ACBH)}
	\end{subfigure}%
	\hfill
	\begin{subfigure}[b]{0.48\linewidth}
		\centering
		\includegraphics[width=0.5\linewidth]{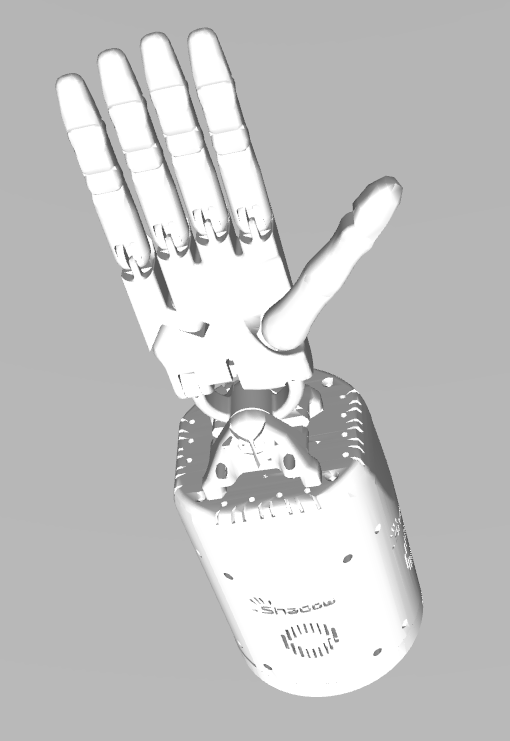}
		\caption{Simulated model of the Shadow Dexterous Hand (SDH)}
	\end{subfigure}
	\caption{Simulation models of the two robotic hands used during analysis~\cite{shadowrobot_urdf, polcz2024posture}.}
	\label{fig:handmodels}
\end{figure}

The morphology of a robotic hand, including the geometric arrangement of its joint
axes, the topology of its actuation system, and the distribution of actuator
authority across its degrees of freedom, determines what manipulation behaviours can
be effectively controlled, and how efficiently a learning algorithm can discover
said behaviours~\cite{pfeifer2006body, hauser2011morphological}. Structural design choices
therefore act as inductive biases on the control and learning problem of robotic
hands~\cite{pfeifer2006body, hauser2011morphological}. Hand designs vary considerably
in how they resolve these morphological trade-offs, ranging from heavily simplified
manipulators to anatomically accurate biomechatronic systems~\cite{salisbury1982articulated,piazza2019century, kashef2020review}.
Advances in computational design and optimization have made it possible to explore
robot morphologies through simulation-based evaluation. These approaches frequently
combine parametric models with performance measures in order to identify designs
that satisfy task-specific constraints. However, many design studies focus
primarily on performance, strictly measured by speed or stability, while the
structural characteristics that may enable adaptability across multiple tasks
remain underexplored.\\
To address this gap, a four-aspect morphological taxonomy is proposed in this work,
namely joint axis geometry, actuator-to-DOF ratio, coupling architecture, and
authority distribution. Two contrasting hand designs are used as case studies to
illustrate how each aspect manifests in standard structural metrics: the Shadow
Dexterous Hand (SDH)~\cite{shadowrobot_techspec,shadowrobot_urdf} and the Anatomically Correct, Biomechatronic
Hand (ACBH)~\cite{tasi2019acbh}. These two hands serve as optimal subjects for
comparison as they sit towards opposing ends of all four aspects. The SDH employs
orthogonal joint axes, a partly underactuated design, sparse explicit coupling, and
a uniform actuator authority distribution. The ACBH employs anatomically oblique
axes, an overactuated design, a branching tendinous network for actuation, and a
specialized authority distribution characteristic of human anatomy. For both hands,
kinematic and actuation parameters are derived directly from their canonical digital
representations (Fig.~\ref{fig:handmodels}). The SDH parameters are obtained from the official URDF, while the
ACBH parameters are obtained from the MuJoCo~\cite{todorov2012mujoco} simulation model developed by Polcz et
al.~\cite{polcz2024posture}, which provides a high-fidelity digital representation
of the physical hand, with some differences arising from the exact physical model
due to factors such as simulation resolution and material characteristics.\\
The objective of this study is twofold. First, to determine which morphological
aspects have measurable structural and functional consequences. Second, to identify
which general design philosophy, i.e.\ mechanical simplification or anatomical
fidelity, is preferable for which aspect. The findings are specific to these two
hands, while the framework can be generalized to any robotic hand for which
kinematic and actuation parameters can be extracted. In addition, the structural
predictions derived from the framework are compared against training outcomes from
reinforcement learning (RL) experiments on both hands, covering two tasks (with
variations within the task's goal, totaling three distinct environments) and three RL
algorithms, in order to assess how well the framework's predictions hold during
training.\\
Section~\ref{sec:methods} develops the mathematical framework and defines the
metrics. Section~\ref{sec:results} introduces the morphological taxonomy and
presents numerical results organized around it. Section~\ref{sec:discussion}
interprets these results in terms of control and learning complexity, while
Section~\ref{sec:conclusion} serves as the conclusion.

\begin{figure}[H]
	\centering
	\includegraphics[width=0.7\linewidth]{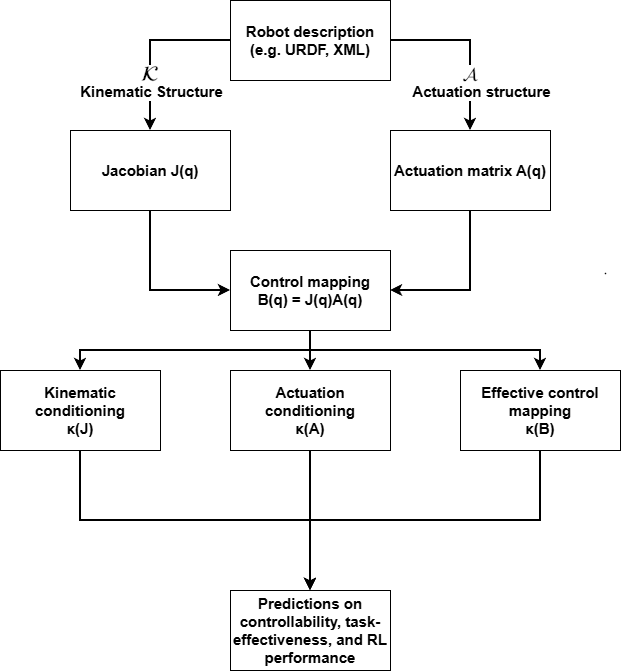}
	\caption{Simplified overview of the analysis pipeline. Each robotic hand is converted from its digital representation to a common structural model $\mathcal{H}=(\mathcal{K},\mathcal{A})$, from which the task Jacobian $J$, the
		actuation matrix $A$, and the effective control mapping $B=JA$ are constructed.}
	\label{fig:diagram}
\end{figure}

\clearpage

\clearpage
\section{Methods}
\label{sec:methods}
The goal of this work is to create a comparison between robotic hand designs at a
structural and functional level. In order to achieve this, a systems-level
formulation is adopted in which morphology and actuation define the space of
achievable coordination patterns, shaping control complexity, learning
effectiveness, and task efficiency~\cite{pfeifer2006body, hauser2011morphological}. The overall structure of the analysis pipeline is summarized in Fig.~\ref{fig:diagram}.

\subsection{Mathematical background}
\label{subsec:maths}
This subsection introduces the mathematical tools used throughout the paper: screw
theory as used to describe joint motion, the Product of Exponentials (PoE)
formulation, singular value decomposition, condition numbers, manipulability, and
motor synergies.

\paragraph{Screw theory and the product of exponentials}
A revolute joint's motion is fully determined by two things: the line in space about which it rotates (i.e. its rotational axis), and the rate at which it turns around that line (i.e. its angular velocity). Screw theory turns this information into a single six-component object, which makes it possible to describe joints whose axes are neither parallel nor intersecting without attaching a coordinate frame to every link.
The instantaneous motion of a rigid body is described by a twist
$\mathcal{V} = (\omega,v) \in \mathbb{R}^6$, where $\omega \in \mathbb{R}^3$ denotes the
angular velocity and $v \in \mathbb{R}^3$ is the linear velocity of the body-fixed point that coincides with the origin of the space frame. For a revolute joint turning at unit rate, the resulting twist denotes the joint's screw axis
$\hat{\xi} = \begin{pmatrix} \omega \\ v \end{pmatrix} \in \mathbb{R}^6$, with
 $\omega$ a unit vector along the axis of rotation and $v = p \times \omega$, where $p \in \mathbb{R}^3$ denotes the position vector of any point on that axis. The choice of $p$ does not affect $v$, since moving $p$ along the axis changes it only by a multiple of $\omega$, the effect of which gets cancelled out by the use of the cross product. The pair $(\omega, v)$ therefore encodes both the direction of the axis and its position in space, independently of any link-attached coordinates. This coordinate-invariance is the main reason the Product of Exponentials (PoE) formulation is preferred here over Denavit--Hartenberg for the structural comparison~\cite{lynch2017modern, murray1994mathematical}.

A twist acts on configurations through the matrix exponential. For a screw axis
$\hat{\xi}$ and joint coordinate $q_i$, the displacement is
$e^{[\hat{\xi}_i] q_i} \in SE(3)$, where $[\hat{\xi}_i]$ is the $4 \times 4$ matrix
form of the screw. Composing the displacements of all joints in a serial chain gives
the forward kinematics as a product of exponentials,
\begin{equation}
T(q) = \left( \prod_{i=1}^{n_f} e^{[\hat{\xi}_i] q_i} \right) T_0,
\end{equation}
with $T_0 \in SE(3)$ the zero-configuration pose of the fingertip frame and $n_f$
the number of joints in the finger. Each screw axis $\hat{\xi}_i$ is expressed in
the space (palm) frame, so the formulation is coordinate-invariant with respect to
the individual links.

\paragraph{Singular value decomposition}
Any real matrix $M \in \mathbb{R}^{a \times b}$ can be subjected to a singular value
decomposition $M = U \Sigma V^\star$, where $U$ and $V$ are orthogonal and $\Sigma$
contains the non-negative singular values $\sigma_1 \geq \sigma_2 \geq \dots \geq 0$
on its diagonal. Singular values represent the stretching factors of all directions
that the matrix acts in, revealing the ``best axes'' geometrically. Singular values
share a mathematical connection with eigenvalues, quantified in this equation:
$\sigma_{i} = \sqrt{\lambda_i (M^TM)}$, for any real matrix $M$.

\paragraph{Condition number}
The condition number
\begin{equation}
\kappa(M) = \frac{\sigma_{\max}(M)}{\sigma_{\min}(M)} \geq 1
\end{equation}
measures the anisotropy of the stretching of matrix $M$. A value near 1 indicates an
almost isotropic mapping that treats all directions comparably, while a large value
indicates that some output directions are reachable only through disproportionately
large inputs, making the mapping sensitive to noise and numerical error along those
directions~\cite{merlet2006jacobian,gosselin1991global,gosselin1992dexterity}.

\paragraph{Manipulability}
The manipulability measure summarizes the kinematic mapping in a single scalar,
\begin{equation}
\omega(q) = \sqrt{\det\!\big(J(q) J(q)^\top\big)},
\end{equation}
which is proportional to the volume of the velocity ellipsoid spanned by the columns
of $J(q)$~\cite{yoshikawa1985manipulability}. Large values indicate that joint
motion produces task-space motion freely in all directions, and values approaching
zero indicate proximity to a singular configuration in which mobility is lost along
one or more directions.

\paragraph{Synergies}
Biological and robotic hands rarely use all of their degrees of freedom
independently. Coordinated patterns of joint motion that recur across many tasks are
called motor synergies. Structurally, a strong synergy corresponds to task-space
behaviour that is well captured by a small number of coordination directions, i.e.\
by a low effective dimensionality of the control mapping ~\cite{santello1998synergies,latash2008synergy,ciocarlie2009subspaces}.

\subsection{Robotic hands as systems}
Let a robotic hand be modelled as a structured mechanical system
\begin{equation}
\mathcal{H} = (\mathcal{K}, \mathcal{A}),
\end{equation}
where $\mathcal{K}$ denotes the hand's kinematic structure, and $\mathcal{A}$ denotes
its actuation structure. The aim of this modeling is to separate the geometrical and
topological properties of the hand from the control strategies applied to it during
operation, resulting in an analysis grounded in physical structure as opposed to
algorithmic performance.

The hand configuration is described by a joint coordinate vector
$q \in \mathbb{R}^n$, with $n$ being the total degrees of freedom (DOFs). Variables
that are relevant during dexterous manipulation tasks, such as fingertip positions
and object orientations, can be represented by a task-space vector
$x \in \mathbb{R}^k$. Then, the forward kinematics of the hand are given by a smooth
mapping
\begin{equation}
x = f(q).
\end{equation}
This formulation can be applied to robotic hands in a general
way~\cite{lynch2017modern}, and differences between the currently analysed ACBH and
SDH are not captured by changes in the task definition, but by the differences in
the structures of $\mathcal{K}$ and $\mathcal{A}$.

\subsection{Kinematic structure}
A hand's kinematic structure $\mathcal{K}$ captures the order of joints, the joint
types, and the geometry of said joints' motion axes. Each finger is modelled as a
serial kinematic chain rooted at a (generally) palm-fixed reference frame. The
fingers' forward kinematics are expressed using the Product of Exponentials (PoE)
formulation~\cite{lynch2017modern, murray1994mathematical}:
\begin{equation}
T(q) = \prod_{i=1}^{n_f} e^{\left(\hat{\xi}_i q_i\right)} T_0,
\end{equation}
with the screw axes $\hat{\xi}_i \in \mathbb{R}^6$, joint
coordinates $q_i$, joint count $n_f$, and zero configuration $T_0$ as defined in Section \ref{sec:methods}.1. In this work, the space-frame screw axes $\hat{\xi}_i$ for the ACBH
are computed using its MuJoCo model's XML kinematic tree. Each body's global
position $p_i$ and orientation $R_i$ in the zero-configuration frame are accumulated
via homogeneous transformations, and for each revolute joint with local axis $a_i$
the global screw axis is calculated by
\begin{equation}
\hat{\xi}_i = \begin{bmatrix} R_i\, a_i \\ p_i \times (R_i\, a_i) \end{bmatrix}.
\end{equation}
For the SDH, screw axes are derived analytically from the published URDF
descriptions. The resulting PoE representations, along with the
Denavit--Hartenberg (D-H) representations used as reference for both hands, are
provided in the Appendix.

This representation has the advantages of being coordinate-invariant and allowing
joint axes of arbitrary orientation to be described without loss of generality.
While D-H parameters are more widely used for implementation and simulation
purposes, the PoE formulation offers a clearer basis for analytical comparisons
between the two hands~\cite{lynch2017modern,murray1994mathematical}.

Kinematic structure may also be examined with the use of singularity analysis, which
can highlight the existence of so-called singular configurations. These
configurations correspond to states in which the system's Jacobian loses rank,
leading to a loss of instantaneous mobility in certain directions within the
task-space. The distribution and frequency of such configurations depend strongly on
joint axis geometry and kinematic redundancy, and can therefore differ significantly
between different hand designs~\cite{muller2019singular}.

Within this framework, the primary difference between the two devices is the
dimensionality and geometry of their kinematic structures. The Shadow Dexterous Hand
employs a reduced number of DOFs per finger, with joint axes that are largely
orthogonal and aligned across fingers. In contrast, the Anatomically Correct,
Biomechatronic Hand includes a higher number of DOFs, accompanied by compound joints
with oblique and non-parallel motion axes derived from human hand
anatomy~\cite{polcz2024posture, valerocuevas2016neuromechanics, santos2006dh}.
These differences directly affect kinematic conditioning and redundancy, as
quantified in Section~\ref{sec:results}.

\subsection{Actuation structure}
The actuation structure $\mathcal{A}$ describes how actuator inputs translate into
joint motions. Let $u \in \mathbb{R}^m$ denote the vector of actuator inputs, with
$m$ being the number of independent actuators. Joint velocities are a direct result
of actuator inputs, with the general mapping
\begin{equation}
\dot{q} = A(q)\,u,
\label{eq:actmap}
\end{equation}
where $A(q) \in \mathbb{R}^{n \times m}$ is the actuation
matrix.

The rank, sparsity, and conditioning of $A(q)$ characterize the control authority and
coordination burden resulting from the hand's actuation design. Underactuated systems
satisfy $m < n$, so that $\mathrm{rank}(A) \leq m < n$, implying that joint motions are mechanically coupled
and that control inputs act along a reduced set of coordination
directions~\cite{dollar2011coupling, laliberte1998underactuated, dollar2010sdm}. Overactuated systems with $m > n$ provide
additional null-space freedom but require the controller to resolve the redundant
degrees of freedom explicitly.

To quantify actuation conditioning, let us consider the condition number
\begin{equation}
\kappa(M) = \frac{\sigma_{\max}(M)}{\sigma_{\min}(M)},
\end{equation}
where $\sigma_{\max}$ and $\sigma_{\min}$ are the largest and smallest singular
values of a generic matrix $M$, respectively~\cite{spong2020robot,
siciliano2009robotics}. A large condition number implies that small changes in
actuator inputs can lead to unexpectedly large or uncontrolled joint motions,
increasing sensitivity to noise and control errors.

\paragraph{SDH actuation matrix}
When each actuator drives either a single joint or a fixed group of joints, the actuation matrix can be derived from the ratios at which actuators drive individual joints or groups of joints. In the case of the SDH, this behaviour holds. For the index,
middle, and ring fingers, the four joint coordinates
$q=(q_{J4},q_{J3},q_{J2},q_{J1})$ are driven by three actuators
$u=(u_{J4},u_{J3},u_{J0})$, where the single actuator $u_{J0}$ drives the PIP and DIP
joints together through a fixed $1{:}1$ coupling within their joint limits. Reading
off these relations gives
\begin{equation}
	A_{\mathrm{SDH,FF}} =
	\begin{pmatrix}
		1 & 0 & 0\\
		0 & 1 & 0\\
		0 & 0 & 1\\
		0 & 0 & 1
	\end{pmatrix}\in\mathbb{R}^{4\times3}.
	\label{eq:a_sdh}
\end{equation}
The rank of $3<4$ expresses the underactuation directly: because the last two rows
are identical, the PIP and DIP coordinates share a single column and cannot be
commanded independently. The little finger adds one independent proximal actuator,
giving $A_{\mathrm{SDH,LF}}\in\mathbb{R}^{5\times4}$ with the same distal coupling,
and the thumb is fully actuated, so $A_{\mathrm{SDH,TH}}=I_5$.

\paragraph{ACBH actuation matrix}
In the case of the ACBH, joints are not driven directly by dedicated actuators but through a tendinous network which branches, recombines, and inserts at multiple "bones", so that each actuator's effect on the joints is configuration dependent~\cite{shirafuji2020branching,valerocuevas2009computational}. The actuation matrix for this system must therefore be derived from the length of each muscle's tendon path as a function of joint angles, with $L_m(\theta)\in\mathbb{R}^{n_m}$ as a collection of $n_m$ muscle-tendon lengths as functions
of the $n_\theta$ joint angles. The rate at which a tendon's length changes per unit joint rotation is the moment arm of that muscle about that joint, so the matrix of
moment arms is the Jacobian of $L_m$ with respect to $\theta$,
\begin{equation}
	R_m(\theta)=\frac{\partial L_m}{\partial\theta}\in\mathbb{R}^{n_m\times n_\theta}.
	\label{eq:Rm}
\end{equation}
Differentiating along a trajectory relates tendon velocities to joint velocities as
$\dot L_m = R_m(\theta)\,\dot\theta$. Inverting this relation gives the joint
velocities produced by a given set of actuator (tendon) velocities; because the system
is overactuated ($n_m>n_\theta$), the inverse is taken via
the Moore--Penrose pseudoinverse, $\dot\theta = R_m(\theta)^{+}\dot L_m$. Comparing
with the general actuation mapping (Eq.~\ref{eq:actmap}), the ACBH actuation matrix at
the reference configuration is therefore
\begin{equation}
	A_{\mathrm{ACBH}}(q)=R_m(\theta_{\mathrm{ref}})^{+}\in\mathbb{R}^{n_\theta\times n_m},
	\label{eq:a_acbh}
\end{equation}
with $u=\dot L_m$. In practice, $R_m(\theta_{\mathrm{ref}})$ is computed numerically by
finite differences over the tendon routing geometry encoded in the MuJoCo XML, and the
branch-level moment arms are aggregated to muscle level through the branch--muscle
incidence matrix $C_{b,m}\in\{0,1\}^{n_b\times n_m}$ of Polcz et al.~\cite{polcz2024posture}.
Verification against the recorded rest-length values agrees to within 1--5\% across all
71 tendon branches.

\subsection{Task-space and control mapping}
Control manipulation tasks generally involve generating actuator inputs that produce
desired task-space motions. Let $x \in \mathbb{R}^k$ denote a task-space variable
such as fingertip position or object pose, and let $q \in \mathbb{R}^n$ denote the
joint configuration. The differential relationship between joint motion and task
motion is given by the Jacobian
\begin{equation}
\dot{x} = J(q)\dot{q},
\end{equation}
where $J(q) \in \mathbb{R}^{k \times n}$ is the task
Jacobian~\cite{lynch2017modern, spong2020robot}. Substituting the actuation mapping
introduced previously yields
\begin{equation} \label{eq:xdot_bu}
\dot{x} = J(q)A(q)u,
\end{equation}
which directly relates actuator inputs to task-space motion. The matrix
\begin{equation} \label{eq:bq}
B(q) = J(q)A(q)
\end{equation}
therefore represents the effective control mapping from actuator inputs to
task-space velocities~\cite{murray1994mathematical}. The structural properties of
$B(q)$ determine how easily the system can generate task motions. In particular, the
rank and conditioning of $B(q)$ characterize both the controllability of task-space
directions and the sensitivity of task execution to actuator coordination.

\subsection{Coordination under bounded actuator effort}
While the full joint space $\mathbb{R}^n$ contains all kinematically possible
configurations, not all configurations are equally reachable under
bounded actuator effort. The set of joint velocities that can be produced under such a limit therefore provides 
a structural description of
the coordination available to a controller at a particular configuration. Since joint
velocities depend linearly on actuator inputs through Eq.~\eqref{eq:actmap}, this set is
the image of a bounded ball of actuator inputs under $A(q)$,
\begin{equation}
	\mathcal{V}(q) = \bigl\{\, A(q)u ~\big|~ u \in \mathbb{R}^m,\
	\|u\| \le \epsilon \,\bigr\} \subset \mathbb{R}^n,
	\label{eq:velset}
\end{equation}
which is an ellipsoid, with its axes pointing along the joint-space directions that the actuators drive most and least effectively, with their lengths measuring how much joint motion a bounded input produces along those directions. Taking the same bounded set of actuator inputs and correlating it to the fingertip motions they produce (instead of the joint motions) can be done by the usage of the effective control mapping $B(q) = J(q)A(q)$ of
Eq.~\eqref{eq:bq}, yielding the corresponding ellipsoid of achievable task-space velocities.

The geometry of these sets carries the most relevant information from the perspective of structural analysis. Spherical, or nearly spherical sets indicate that bounded actuator effort produces task-space motion of
comparable magnitude in every direction, whereas elongated sets indicate the existence
of directions along which motion can be produced only by exerting disproportionate
effort. The ratio between the longest and the shortest semi-axis is exactly the
condition number of the corresponding matrix. Directions associated with the
smallest singular values are those which a controller, or a learning algorithm exploring
in actuator space, can exploit only at significant cost, and where such directions are
strongly disfavoured, the coordination patterns that avoid them become mechanically
preferred, which is the structural analogue of the motor synergies introduced in
subsection~\ref{subsec:maths}~\cite{santello1998synergies,latash2008synergy,ciocarlie2009subspaces}.

\begin{figure}[!htbp]
	\centering
	\begin{subfigure}[b]{0.48\linewidth}
		\centering
		\includegraphics[width=\linewidth]{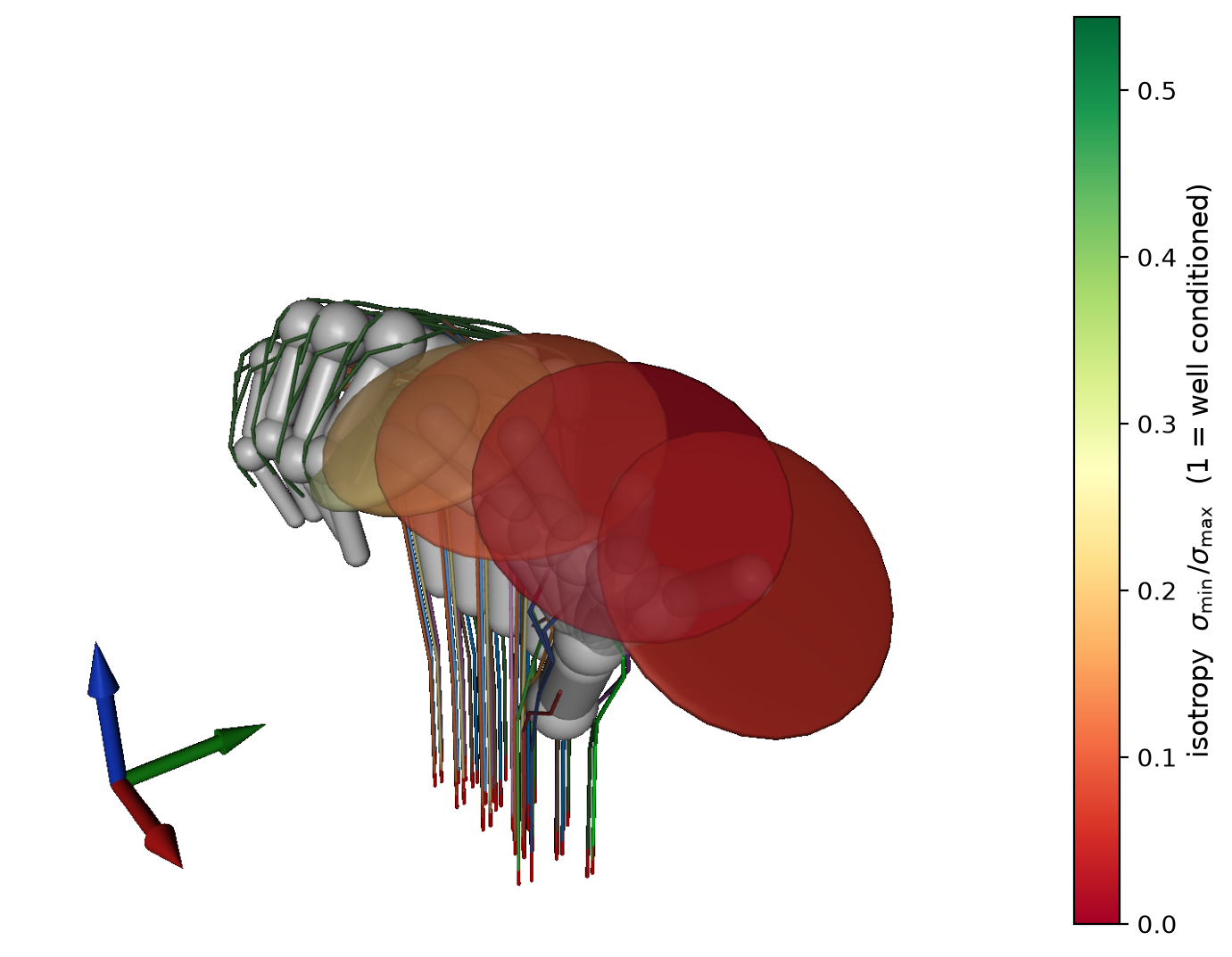}
		\caption{}
	\end{subfigure}%
	\hfill
	\begin{subfigure}[b]{0.48\linewidth}
		\centering
		\includegraphics[width=\linewidth]{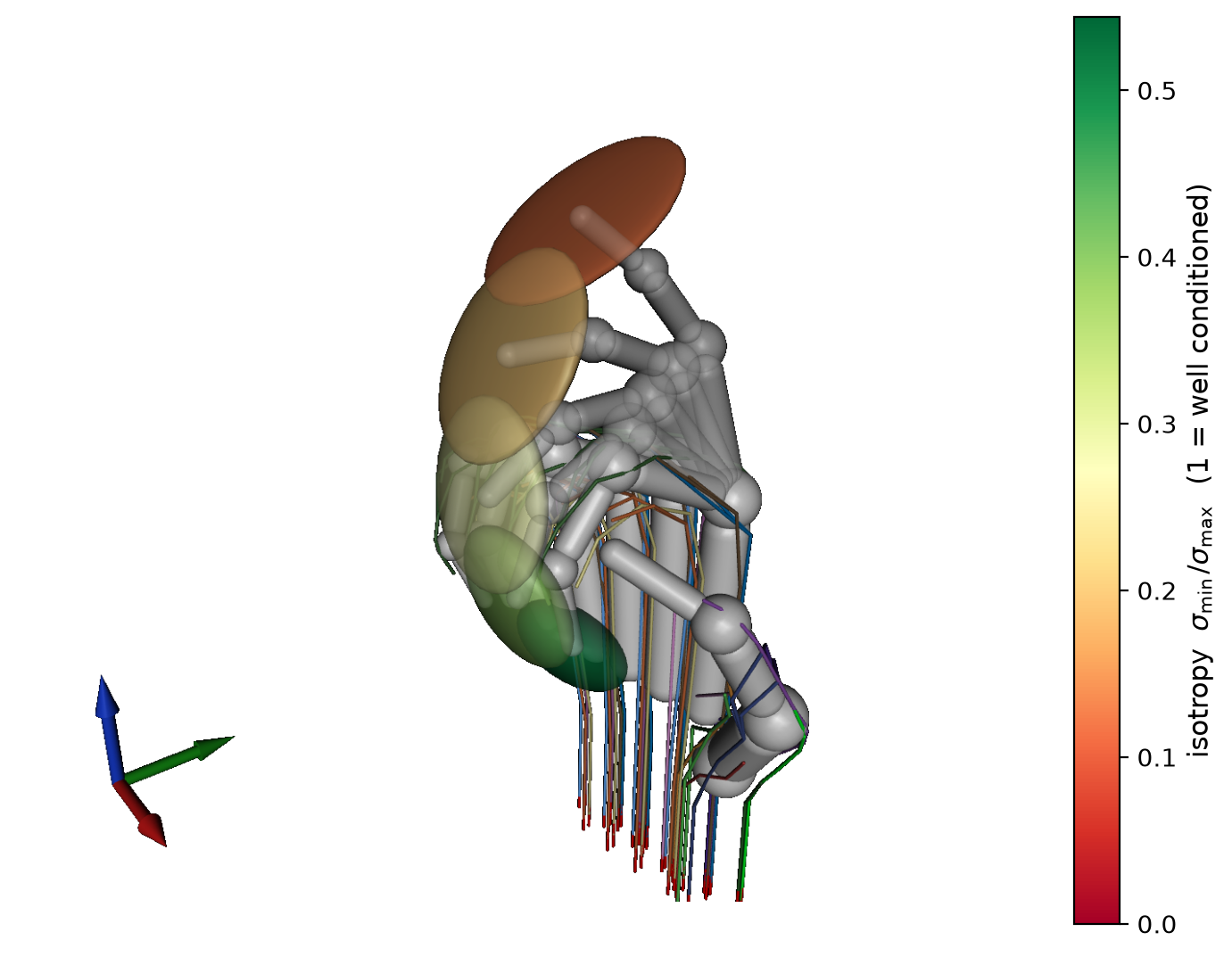}
		\caption{}
	\end{subfigure}
	\caption{Visualization of the configuration dependence of the kinematic conditioning of the ACBH's thumb (a) and index finger (b), using overlaid positions. At each position, the velocity ellipsoid of the task Jacobian $J(q)$ is drawn at the fingertip. Ellipsoids are coloured by their isotropy $\sigma_{\min}/\sigma_{\max}$, from green (well conditioned) towards red (ill-conditioned). The semi-axes of each ellipsoid are proportional to the singular values of the corresponding mapping, so a long axis denotes a direction in which task-space motion is produced most readily. Poor conditioning is indicated not by the ellipsoids' overall sizes, but by their elongation: a near-spherical ellipsoid denotes a mapping that treats all directions comparably, while an elongated one denotes a large ratio between the most and least accessible directions.
	The ellipsoids are evaluated at the configurations shown, whereas the metrics reported in Table \ref{tab:per_finger} are evaluated at the mid-range reference configuration $q_{ref}$. The orientation of the space frame is denoted with the x, y and z axes drawn in red, green and blue, respectively. }
	\label{fig:kappa_J}
\end{figure}

\begin{figure}[!htbp]
	\centering
	\begin{subfigure}[b]{0.48\linewidth}
		\centering
		\includegraphics[width=\linewidth]{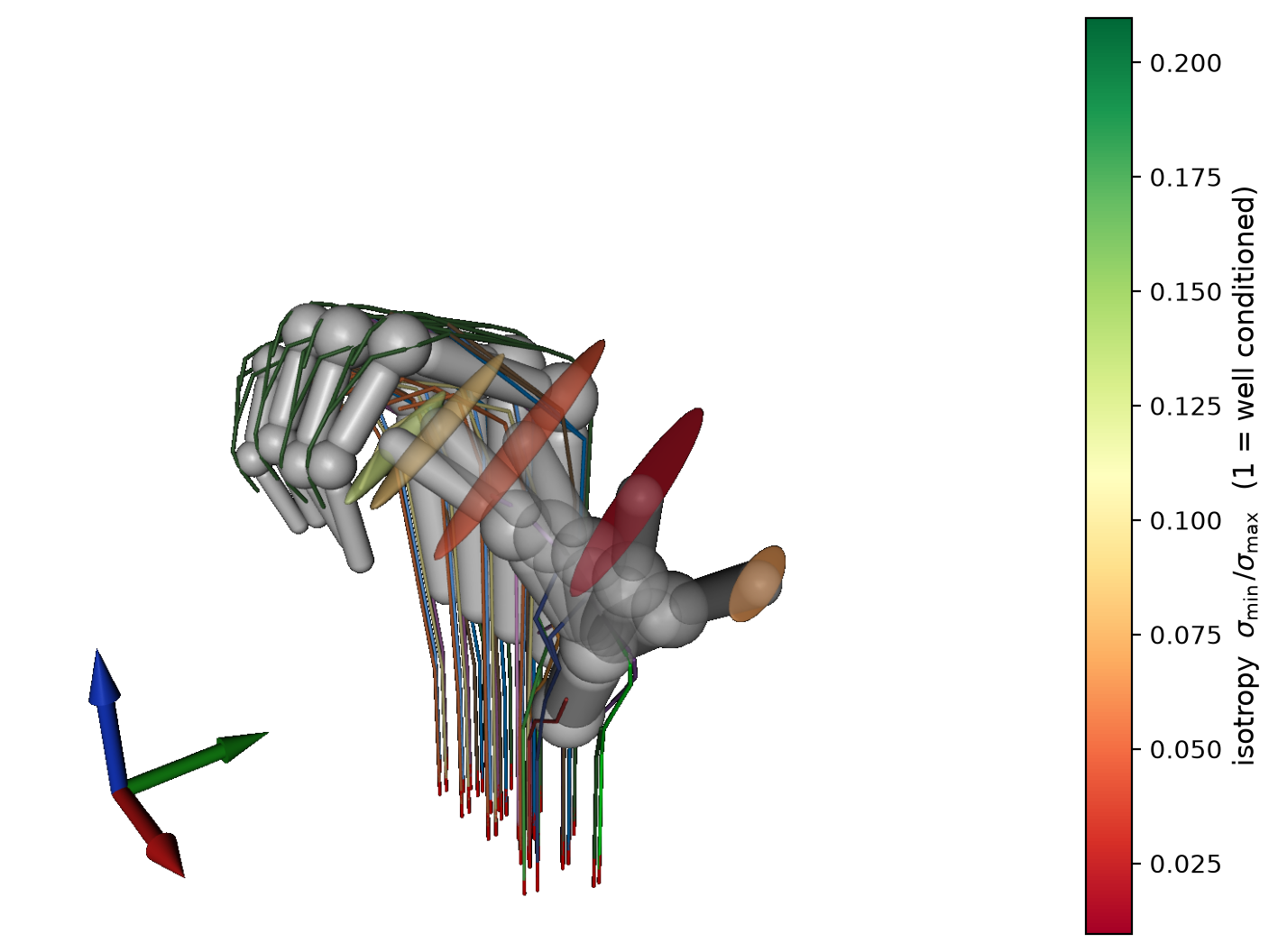}
		\caption{}
	\end{subfigure}%
	\hfill
	\begin{subfigure}[b]{0.48\linewidth}
		\centering
		\includegraphics[width=\linewidth]{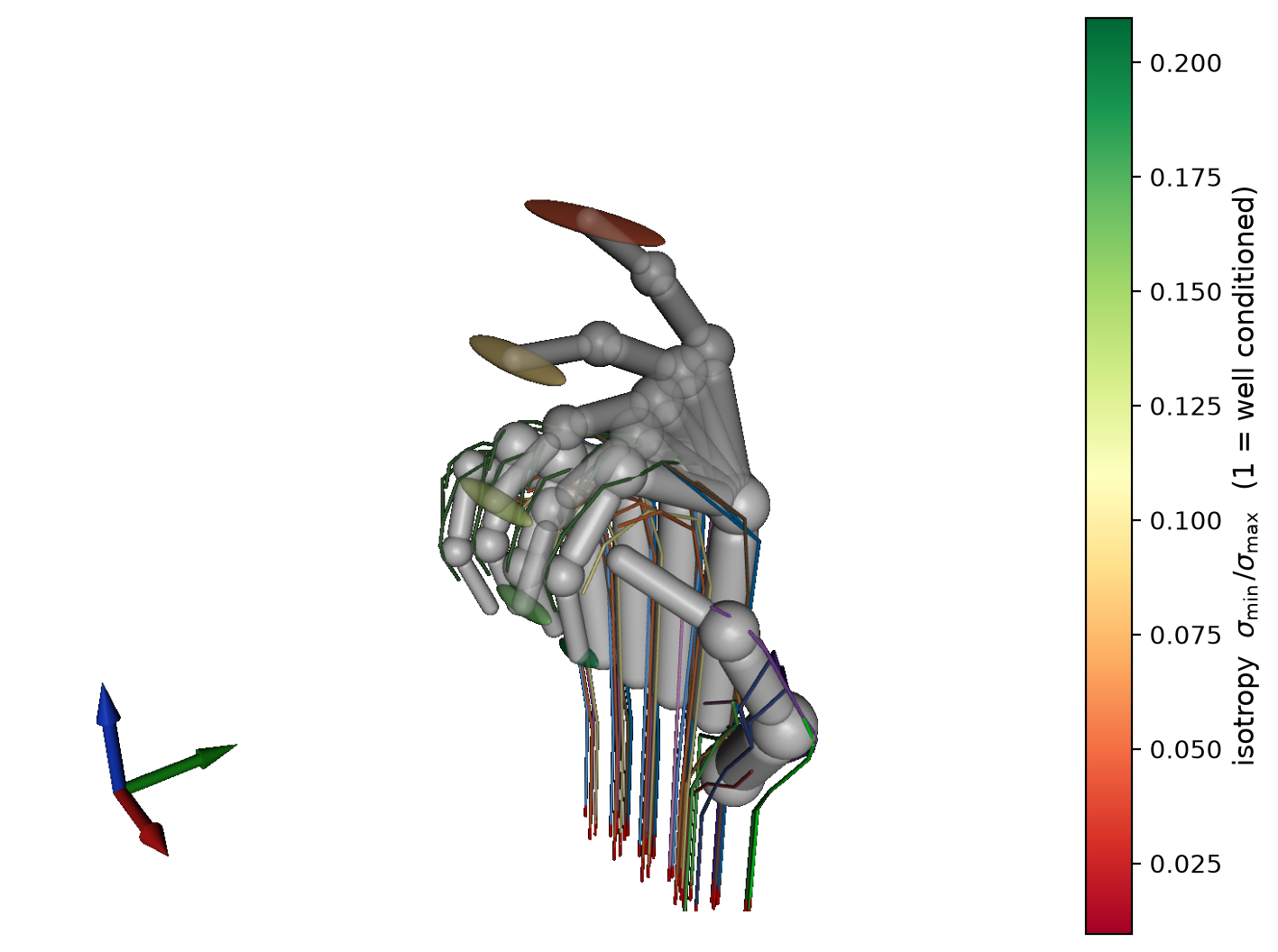}
		\caption{}
	\end{subfigure}
	\caption{Visualization of the configuration dependence of the effective control conditioning of the ACBH's thumb (a) and index finger (b), using overlaid positions. At each position, the velocity ellipsoid of the effective control mapping $B(q)$ is drawn at the fingertip. The semi-axes of each ellipsoid are proportional to the singular values of the corresponding mapping, so a long axis denotes a direction in which task-space motion is produced most readily. Poor conditioning is indicated not by the ellipsoids' overall sizes, but by their elongation: a near-spherical ellipsoid denotes a mapping that treats all directions comparably, while an elongated one denotes a large ratio between the most and least accessible directions.
	The ellipsoids are evaluated at the configurations shown, whereas the metrics reported in Table \ref{tab:per_finger} are evaluated at the mid-range reference configuration $q_{ref}$. The orientation of the space frame is denoted with the x, y and z axes drawn in red, green and blue, respectively.}
	\label{fig:kappa_B}
\end{figure}

Both $A(q)$ and $J(q)$ vary across the workspace, so these sets are configuration
dependent. Figures~\ref{fig:kappa_J} and~\ref{fig:kappa_B} illustrate this dependence for
the ACBH by drawing the corresponding ellipsoids at a range of configurations, whereas
the metrics reported in Section~\ref{sec:results} are evaluated at the mid-range
reference configuration $q_{\text{ref}}$.

\subsection{Structural metrics}
To evaluate the structural characteristics of the robot configurations, a set of
structural metrics is applied. The selected metrics focus on three primary structural
attributes: conditioning, redundancy, and manipulability.

\subsubsection{Kinematic conditioning}
Even when the rank of $J(q)$ is sufficient to achieve a desired task motion, control
may remain difficult if the kinematic mapping from joint velocities to task-space
velocities is poorly conditioned. Conditioning can be measured by
\begin{equation}
	\kappa(J)\big|_{q_{\mathrm{ref}}}=\frac{\sigma_{\max}\!\bigl(J(q_{\mathrm{ref}})\bigr)}
	{\sigma_{\min}\!\bigl(J(q_{\mathrm{ref}})\bigr)},
	\label{eq:kJ}
\end{equation}
where $\sigma_{\max}$ and $\sigma_{\min}$ are the largest and smallest singular values
of $J(q)$, respectively~\cite{spong2020robot,siciliano2009robotics}. Large
conditioning numbers indicate the existence of certain task-space directions that
require disproportionately large joint velocities to produce, increasing sensitivity to
joint noise and configuration errors. Well-conditioned Jacobians produce more uniform
task-space coverage per unit joint motion, which has direct implications for the
isotropy of exploration in learning-based manipulation. Unless stated otherwise, all
metrics are evaluated at a reference configuration $q_{\text{ref}}$ defined as the
midpoint of each joint's permitted range of motion:
\begin{equation} \label{eq:qref}
q_{\text{ref},i} = \tfrac{1}{2}(q_{i,\min} + q_{i,\max}).
\end{equation}

\clearpage

\subsubsection{Redundancy}
Redundancy in this context describes the presence of additional structural elements or
alternative pathways within a system that allow the structure to maintain
functionality under varying conditions. While mechanical redundancy can increase the
flexibility of the hands, it can affect the overall performance negatively in
manipulation tasks due to increasing control complexity and coordination burden.
Redundancy as a metric is entirely task-specific and can be simply calculated as
\begin{equation}
r = n-k,
\end{equation}
where $n$ is the hand's number of degrees of freedom and $k$ is the specific task's
dimensionality~\cite{lynch2017modern}.

\subsubsection{Manipulability}
Manipulability describes the ability of a mechanical system to generate motion in
different directions through coordinated actuation of its
joints~\cite{yoshikawa1985manipulability}. In robotic systems, manipulability reflects
how effectively joint inputs can produce movement of the mechanism within its
operational space. It can be calculated using the system's Jacobian:
\begin{equation}
\omega(q_{\mathrm{ref}})=\sqrt{\det\!\bigl(J(q_{\mathrm{ref}})J(q_{\mathrm{ref}})^{\top}\bigr)}.
\label{eq:omega}
\end{equation}
$\omega(q)$ quantifies the local ability of the hand to generate task-space motions in
different directions through joint actuation. Configurations with larger values of
$\omega(q)$ allow the system to produce motion more uniformly throughout different
task-space directions, indicating that the joint axes are arranged in a way that
avoids kinematic singularities or strongly constrained motion directions. Conversely,
low manipulability values imply that certain directions in task-space require
disproportionately large joint motions. For dexterous manipulation, high
manipulability is desirable when tasks require fine adjustments of fingertip position
or orientation~\cite{yoshikawa1985manipulability, lynch2017modern}. Manipulability,
therefore, captures one aspect of kinematic structure, being complemented by
$\kappa(J)$ and $\kappa(B)$. For multi-finger tasks, $\omega(q)$ is evaluated per
finger only, to keep results meaningful, which cannot be ensured while using the
block-diagonal Jacobian that is calculated for combined task descriptions.

\subsection{Effective control-mapping conditioning}
While $\kappa(J)$ captures the coordination burden imposed by joint geometry alone,
the burden actually faced by a controller depends on how actuator inputs propagate to
task-space velocities. Substituting the actuation mapping into the differential
kinematics recovers the effective control mapping of Eqs. \ref{eq:xdot_bu} and \ref{eq:bq}, $\dot{x} = J(q)A(q)u = B(q)u$,
where $B(q) \in \mathbb{R}^{k \times m}$ is the effective control mapping defined in
this section. Its condition number
\begin{equation}
\kappa(B)\big|_{q_{\mathrm{ref}}}=\frac{\sigma_{\max}\!\bigl(B(q_{\mathrm{ref}})\bigr)}
{\sigma_{\min}\!\bigl(B(q_{\mathrm{ref}})\bigr)}
\label{eq:kB}
\end{equation}
quantifies the directional uniformity of the actuator-to-task mapping. A comparison
between $\kappa(B)$ and $\kappa(J)$ reveals the extent to which a system's actuation
architecture redistributes (or concentrates) the kinematic coordination burden.

\clearpage

\paragraph{Actuation conditioning}
While the metrics $\kappa(J)$ and $\kappa(B)$ provide information about the mapping at its two respective ends, kinematic input and task-space output, the actuation stage between the two is left implicitly. Applying the same condition number to the actuation matrix isolates that stage, and measures the directional uniformity with which actuator inputs are distributed
across joint coordinates, independently of the task:
\begin{equation}
	\kappa(A)\big|_{q_{\mathrm{ref}}}=\frac{\sigma_{\max}(A)}{\sigma_{\min}(A)}.
	\label{eq:kA}
\end{equation}
Together, the three metrics decompose the coordination burden along the two stages through which an actuator input reaches the task space. Actuator inputs first produce joint velocities, and joint velocities then produce task-space velocities. The conditioning of the first stage only is measured by $\kappa(A)$, that of the second stage only is measured by $\kappa(J)$, and the combined conditioning of the composition of the two is measured by $\kappa(B)$. Reading them side by side can show where a bottleneck originates, whether it is the system's geometry, actuation, or the interaction of the two. For the SDH, $A$ is a
constant selection matrix and $\kappa(A)$ is therefore configuration-independent; for
the ACBH, $A=R_m^{+}$, so $\kappa(A)=\kappa(R_m)$, the conditioning of the moment-arm
Jacobian itself.

\subsection{RL algorithms used for empirical evaluation}
During this work, three reinforcement learning algorithms were used to generate
empirical training data to be compared with the structural metrics defined above~\cite{raffin2021sb3}.

\paragraph{Proximal Policy Optimization}
Proximal Policy Optimization (PPO) is an RL algorithm based on Trust Region Policy
Optimization (TRPO), expanding on it to make it simpler and more computationally
efficient. PPO is on-policy, meaning that learning is based on the data generated by
the same policy that is being improved, effectively making the agent learn from its
own actions. It is also an actor-critic method, using an actor (or policy network)
that is responsible for outputting parameters for the next actions, and a critic (or
value network) that estimates the ``goodness'' of the current and each possible future
state. The output of these two networks is a probability distribution from which the
next action is sampled, based on the estimated values of future states. PPO optimizes
the expected advantage of actions, but clips the updates for the purposes of
preserving stability. Concretely, PPO constrains each update by clipping the probability ratio between the
new and old policies to a trust region, which prevents destructively large steps and
is the main source of its stability. The advantage estimates that drive the update are
formed by Generalized Advantage Estimation, which trades bias against variance through
a single smoothing parameter. PPO is widely used in continuous control tasks because it has
proven itself to be a stable, sample-efficient, and relatively easy-to-use
algorithm~\cite{schulman2017ppo}.

\paragraph{Deep Deterministic Policy Gradient + Hindsight Experience Replay}
Deep Deterministic Policy Gradient (DDPG) is an off-policy, model-free actor-critic
algorithm. It maintains a similar actor-critic structure as PPO, with the notable
difference in the roles: the actor is responsible for outputting a specific
(deterministic) action, and the critic is responsible for estimating $Q(s, a)$, the
Q-value representing the value of a specific action in a given state. DDPG is
off-policy, meaning it uses data from previous runs stored in a buffer, making it
highly data efficient. However, sparse-reward environments pose a challenge, since
successful episodes are generally rare, especially in high-dimensional action spaces.
Hindsight Experience Replay (HER) is a goal-relabelling strategy that addresses this problem.
After each episode, HER relabels past transitions with alternative goals, turning
failures into useful data points, improving exploration and learning efficiency in
scarce reward scenarios~\cite{lillicrap2015ddpg, andrychowicz2017her}.

\paragraph{Truncated Quantile Critic + Hindsight Experience Replay}
Truncated Quantile Critic (TQC) is another off-policy actor-critic algorithm that
works by modeling the full distribution of expected future returns across a collection
of multiple critics. Instead of outputting a simple average from these returns, each
critic predicts a set of distinct quantile locations to combat environmental
uncertainty. During the training step, TQC collects these predictions into a pool,
sorts them, and drops a predetermined number of the highest values from the
distribution. This truncation filters out overly optimistic outliers and thereby
controls overestimation bias, which is a recurring problem for other value-based RL
algorithms, especially during long training runs. The remaining quantiles are then
averaged to compute a stable target for updating the actor network~\cite{kuznetsov2020tqc}.

\subsection{RL training environments}
Training experiments were performed on both hands across three distinct environments, described formally below~\cite{plappert2018multigoal,openai2020dexterous}. Algorithm choice was matched to task difficulty: the Reach
task was trained with the on-policy PPO, while the sparse-reward block manipulation tasks required the sample efficiency of off-policy methods combined with HER. BlockRotateXYZ, the most demanding environment, was trained with both DDPG+HER and TQC+HER to compare a baseline off-policy method against a more capable one on the hardest task.

\paragraph{Reach}
The Reach task requires the hand to move its end-effectors to a specified target
location in three-dimensional space. More precisely, for each episode, the thumb and a
randomly selected long finger must reach two points in space that are closely located,
resulting in a ``pinching'' gesture.

\paragraph{Block Manipulate -- Rotate Z}
The BlockRotateZ task requires the hand to achieve stable fingertip contact with a
block object and apply coordinated force to rotate it along the vertical axis, in order
to reach a predefined target orientation. This variant randomly shuffles the cube only
along the Z axis for each episode.

\paragraph{Block Manipulate -- Rotate XYZ}
The BlockRotateXYZ task requires rotation of a block object along all three spatial
dimensions and represents the most demanding environment in this study. This
environment was slightly modified for the training runs demonstrated in this study: the rotation success threshold was changed to $0.4$ radians
from the canonical $0.1$ radians.

\clearpage
\section{Results}
\label{sec:results}
This section introduces the results of the four-aspect morphological framework that
distinguishes the two hands under study, reports per-finger structural metrics
organized around this framework, extends the analysis to two manipulation tasks, and
presents an investigation of the most extreme finding. All metrics are evaluated at the
mid-range reference configuration $q_{\text{ref}}$ (Eq.~\ref{eq:qref}).
Table~\ref{tab:fw} characterizes the hands generally with respect to each morphological aspect and the main metric(s) used to capture the structural consequences of these characterizations.
\begin{table}[H]
\centering
\small
\renewcommand{\arraystretch}{1.2}
\begin{tabular}{p{3.0cm}p{2.9cm}p{2.9cm}p{2.0cm}}
\toprule
\textbf{Morphological aspect} & \textbf{SDH} & \textbf{ACBH} & \textbf{Affects} \\
\midrule
\textbf{Joint axis geometry} &
  Orthogonally aligned &
  Anatomically oblique &
  $\kappa(J)$ \\
\textbf{Actuator-to-DOF ratio} ($m/n$) &
  $0.82$ (underactuated) &
  $1.50$ (overactuated) &
  $r$, $A$ shape \\
\textbf{Coupling architecture} &
  Sparse, explicit &
  Branching tendon network &
  $\kappa(B) - \kappa(J)$ \\
\textbf{Authority distribution} &
  Uniform across joints &
  Specialized &
  $\kappa(B)$ outliers \\
\bottomrule
\end{tabular}
\caption{Morphological aspects of comparison. Each row defines a structural design
aspect along which robotic hands can vary. The rightmost column includes the metrics
mainly associated with each aspect.}
\label{tab:fw}
\end{table}

\subsection{Per-finger metrics}
Table~\ref{tab:per_finger} summarizes the metrics for each finger of both hands under
single-fingertip positioning tasks, i.e.\ positioning the fingertips to a specified
location in 3-D space ($k = 3$).

\begin{table}[H]
\centering
\begin{tabular}{llcccccc}
\toprule
Hand & Finger & $n$ & $m$ & $r$ & $\kappa(J)$ & $\kappa(A)$ & $\kappa(B)$ \\
\midrule
SDH  & Index/Middle/Ring & 4 & 3 & 1 & 3.84 & 1.41 & 4.25 \\
SDH  & Little (LF)       & 5 & 4 & 2 & 4.10 & 1.41 & 3.69 \\
SDH  & Thumb (TH)        & 5 & 5 & 2 & 9.45 & 1.00 & 9.45 \\
\midrule
ACBH & Index (F2)        & 5 & 7  & 2 & 9.55 & 6.39 & 6.57 \\
ACBH & Middle (F3)       & 5 & 7  & 2 & 6.90 & 6.48 & 4.85 \\
ACBH & Ring (F4)         & 5 & 7  & 2 & 6.28 & 6.52 & 4.41 \\
ACBH & Little (F5)       & 5 & 7  & 2 & 5.77 & 6.53 & 4.10 \\
ACBH & Thumb (T1)        & 6 & 11 & 3 & 4.76 & 30.74 & 20.03 \\
\bottomrule
\end{tabular}
\caption{Per-finger metrics at $q_{\text{ref}}$. $r = n - k$ with $k = 3$.}
\label{tab:per_finger}
\end{table}

\paragraph{Joint axis geometry}
The orthogonal axis arrangement of the SDH and the oblique axis arrangement of the ACBH
have similar consequences for $\kappa(J)$ at the long fingers. The SDH achieves
$\kappa(J) = 3.84$ on the index (FF), middle (MF), and ring (RF) fingers, which is
lower than any of the ACBH long fingers (range $5.77$ to $9.55$). The two design
philosophies are therefore comparable at this scale, with the SDH showing slightly
better kinematic conditioning. The exception is the thumb, where the ACBH's
anatomically grounded joint representation yields $\kappa(J) = 4.76$, against the SDH's
$\kappa(J) = 9.45$. Therefore, anatomical joint geometry yields a clear kinematic
advantage for the thumb.

\paragraph{Actuator-to-DOF ratio}
The SDH operates at $m/n < 1$ (underactuated), which gives each long finger $r = 1$,
and $r = 2$ for the little finger and the thumb. The ACBH operates at $m/n > 1$
(overactuated, with $7$ to $11$ muscles per finger driving $5$ or $6$ DOFs), which
gives every finger $r \geq 2$. A higher $m$ also widens the column count of $B$,
providing the controller with more degrees of freedom along which to satisfy a given
task.

\paragraph{Coupling architecture}
This is the aspect where the two hands diverge most clearly. For the SDH long fingers,
the J0 coupling (i.e., the actuator that drives the PIP and DIP joints together) slightly worsens the conditioning ($\kappa(B) = 4.25$ versus
$\kappa(J) = 3.84$). The ACBH's branching tendon network has the opposite effect on its
long fingers, with $\kappa(B) < \kappa(J)$ in all four cases. The natural expectation is that added mechanical complexity should raise the coordination burden on the controller. The ACBH long fingers exhibit the opposite, and the reason is that complexity of this particular
kind does not concentrate authority but distributes it. By spreading each joint's
task-space authority across several muscles whose contributions are directionally
complementary, the branching network lowers the effective conditioning rather than
raising it. The network embeds a favourable distribution of authority that a simpler mechanism would leave to the controller to discover.

\paragraph{Authority distribution}
The thumb is an outlier, with ACBH $\kappa(B) = 20.03$ against the SDH thumb's
$\kappa(B) = 9.45$. This is the only finger in either hand for which the actuation
network considerably degrades the underlying kinematic conditioning
($\kappa(J) = 4.76$ but $\kappa(B) = 20.03$). The cause is a non-uniform distribution
of actuator authority across the thumb's joints, which is investigated in this
section's final subsection.

\paragraph{Actuation conditioning}
Separating the actuation stage with $\kappa(A)$ makes it possible to tell from where each
hand's coordination burden arises. For the SDH, the actuation has little significance. Its $\kappa(A)$ never exceeds $\sqrt2\approx1.41$ at any finger, because its actuation matrix is very nearly an isometry. As a result, the SDH's $\kappa(B)$
follows its $\kappa(J)$ closely everywhere, which is the structural reason its
behaviour is so uniform across fingers. The ACBH behaves in the opposite way, especially when considering the thumb: there $\kappa(A)\approx30$, larger even than the composite
$\kappa(B)=20.03$, while its kinematic conditioning $\kappa(J)=4.76$ is among the best in the derived values. The bottleneck originates in the actuation stage, partially mitigated by the kinematics, since $\kappa(B) = 20.03$ falls below $\kappa(A) = 30.74$.  At the long fingers the picture is more subtle: despite
$\kappa(A)>1$, the ACBH reaches $\kappa(B)<\kappa(J)$ on all four, which the magnitude
of $\kappa(A)$ alone cannot explain. A poorly conditioned actuation matrix can still
improve the composite mapping when its authority is directionally aligned with the
kinematics, so the benefit of the branching network at the long fingers is thanks to alignment, not conditioning magnitude.

\subsection{Task 1 -- Precision pinch}
The precision pinch task requires two fingers (thumb and index in this case) to achieve
stable contact with a small object, with task-space dimensionality $k = 6$ (three
position coordinates per fingertip). Table~\ref{tab:tasks} reports the structural
metrics for this task. The ACBH pinch engages 11 joints and 18 muscles ($r = 5$) with
$\kappa(J) = 9.55$ and $\kappa(B) = 20.03$. The SDH pinch uses $9$ joints and $8$
actuators ($r = 3$) with $\kappa(J) = 9.45$ and $\kappa(B) = 9.45$. The kinematic
conditioning of the two pinches is therefore almost identical ($\kappa(J)$), with their
effective actuation conditioning ($\kappa(B)$) diverging by more than a factor of two.
This is a direct consequence of the ACBH's thumb conditioning dominating any combined
task involving it.

\subsection{Task 2 -- Power grasp}
The power grasp task requires all five fingers to envelop an object, with $k = 15$
(three translational coordinates per finger). As visible in Table~\ref{tab:tasks}, the
ACBH engages all 26 joints and $39$ muscles ($r = 11$) with $\kappa(J) = 9.55$ and
$\kappa(B) = 20.03$. The SDH engages $22$ joints and $18$ actuators ($r = 7$) with
$\kappa(J) = 9.70$ and $\kappa(B) = 10.01$. The pattern from the pinch task persists at
this scale, in some cases with exact numerical equivalence. The long-finger advantage of the ACBH's coupling architecture does not
propagate to the combined system, since thumb specialization once again sets the
bottleneck.

\begin{table}[H]
\centering
\begin{tabular}{llcccccc}
\toprule
Hand & Task & $n$ & $m$ & $r$ & $\kappa(J)$ & $\kappa(A)$ & $\kappa(B)$ \\
\midrule
ACBH & Pinch (T1 + F2)         & 11 & 18 &  5 & 9.55 & 30.74 & 20.03 \\
SDH  & Pinch (TH + FF)         &  9 &  8 &  3 & 9.45 & 1.41 & 9.45 \\
\midrule
ACBH & Power grasp (5 fingers) & 26 & 39 & 11 & 9.55 & 30.74 & 20.03 \\
SDH  & Power grasp (5 fingers) & 22 & 18 &  7 & 9.70 & 1.41 & 10.01 \\
\bottomrule
\end{tabular}
\caption{Task-level structural metrics. $r = n - k$ with $k = 6$ (pinch) or $k = 15$
(power grasp).}
\label{tab:tasks}
\end{table}

\subsection{Authority distribution in the ACBH thumb}
The thumb's $\kappa(B) = 20.03$ is the most extreme finding of the analysis and needs
further investigation. Singular value analysis of the $11 \times 6$ thumb moment arm
Jacobian $R_{m,T1}(\theta_{\text{ref}})$ shows that it is full-rank, with six non-zero
singular values, so the high $\kappa(B)$ does not arise from rank deficiency. The cause
is instead visible in the column-norm distribution of $R_{m,T1}$
(Table~\ref{tab:thumb_authority}), which measures how much actuator authority each
joint receives from the thumb muscle group collectively.

\begin{table}[H]
\centering
\begin{tabular}{lcc}
\toprule
Joint & $\|R_{m,T1}[:,j]\|$ & Muscles acting \\
\midrule
TCMC\_Yaw   & 0.046 & 11/11 \\
TCMC\_Pitch & 0.037 & 11/11 \\
MCP\_Roll   & 0.025 &  8/11 \\
MCP\_Yaw    & 0.017 &  8/11 \\
MCP\_Pitch  & 0.0022 & 7/11 \\
IP\_Pitch   & 0.011 &  2/11 \\
\bottomrule
\end{tabular}
\caption{ACBH thumb joint-level actuator authority at $q_{\text{ref}}$, with joints ordered from proximal to distal.}
\label{tab:thumb_authority}
\end{table}

The disparity in joint-level authority, combined with the IP joint's reliance on only
two of the eleven thumb muscles, produces a $\sigma_{\min}(R_{m,T1})$ that is roughly
$30\times$ smaller than $\sigma_{\max}$, which propagates directly into $\kappa(B)$.

This pattern is anatomically consistent. The thenar muscle group is optimized for thumb
opposition, i.e.\ a coordinated abduction and rotation that brings the thumb's pad
opposite the other fingertips~\cite{kapandji2007physiology,inouye2014tendon,valerocuevas2003thumb}. The high
$\kappa(B)$ of the ACBH's thumb is therefore not a modelling error but a manifestation
of authority distribution in an extreme form, since the actuator authority distribution
is highly anisotropic.

This is a key illustration of how morphology acts as a structural inductive
bias~\cite{pfeifer2006body}. The effect of authority distribution is invisible to
kinematic analysis alone, since $\kappa(J) = 4.76$ for the ACBH thumb is one of the best
in this study. It emerges only when $A(q)$ is explicitly constructed.

\subsection{Empirical RL training results}
Training curves for all experiments are provided as Supplementary Material in the Appendix
(Supplementary Figures~\ref{fig:reach_success_rate}--\ref{fig:rotate_xyz_tqc_success_rate}).
All reported success rates are smoothed using an exponential moving average with weight
$\alpha = 0.6$.

\paragraph{Reach (PPO)}
The SDH converged at approximately $11$ million timesteps (${\approx}$~6~h) with a rollout success rate of
$0.75$, while the ACBH converged at
approximately $25$ million timesteps (${\approx}$~19.5~h) with a success rate of $0.65$ (Supplementary
Fig.~\ref{fig:reach_success_rate}).

\paragraph{BlockRotateZ (DDPG+HER)}
At approximately $16$ million timesteps (${\approx}$~23.5~h), the ACBH reached a smoothed ($\alpha = 0.6$)
success rate of $0.692$, while the SDH reached $0.665$ using the same number of timesteps (${\approx}$~9~h)  (Supplementary
Fig.~\ref{fig:rotate_z_success_rate}). The two hands perform comparably over the course
of training. The ACBH's success rate curve is considerably smoother across the full
training run, while the SDH exhibits semi-periodic dips approximately every 2 million
timesteps.

\paragraph{BlockRotateXYZ (DDPG+HER)}
Neither hand achieved a meaningful success rate over the course of training, with both
remaining consistently below $0.1$ (Supplementary
Fig.~\ref{fig:rotate_xyz_ddpg_success_rate}).

\paragraph{BlockRotateXYZ (TQC+HER)}
Under TQC+HER, the SDH achieved a success rate of ${\sim}0.76$ after ${\sim}23$ million
timesteps (${\approx}$~34~h), while the ACBH achieved a ${\sim}0.25$ success rate during the same amount of
timesteps (${\approx}$~73~h) (Supplementary Fig.~\ref{fig:rotate_xyz_tqc_success_rate}).

\clearpage
\section{Discussion}
\label{sec:discussion}
The structural metrics reported in Section~\ref{sec:results} allow each morphological
aspect to be tied to a specific pattern. The benefits of each hand's design choices
depend on factors such as the task being performed, or whether the metric is at the
kinematic or the actuation level. The non-uniformity of these findings is the central
observation of the analysis.

\subsection{Structural patterns across the morphological aspects}

\paragraph{Joint axis geometry}
Anatomically oblique axes do not provide a general kinematic advantage over
orthogonally aligned axes. At the long fingers, the SDH is better conditioned than any
ACBH long finger ($3.84$ against $5.77$--$9.55$). The kinematic advantage of anatomical
fidelity in this case is concentrated at the thumb, where the ACBH's conditioning
reaches $\kappa(J) = 4.76$ against the SDH's $9.45$. This pattern reflects the biological
thumb's unique anatomy, since its kinematic complexity is what allows opposition, and
the SDH's simplified four-joint sequential chain implementation inherits a more
constrained task-space coverage that no choice of joint angles can fully compensate for.
For the long fingers, by contrast, the straightforward MCP--PIP--DIP geometry is
well-suited to fingertip positioning regardless of axis orientation.

\paragraph{Actuator-to-DOF ratio}
A higher $m/n$ provides redundancy but does not improve conditioning by itself. The
ACBH's $m/n = 1.50$ results in $r = 2$ to $3$ on every finger, while the SDH's
$m/n = 0.82$ gives $r = 1$ to $2$. Redundancy is useful for tasks beyond fingertip
positioning, such as contact force modulation, collision avoidance, or internal posture
preservation, but it does not directly reduce $\kappa(J)$ or $\kappa(B)$. What
overactuation does provide is a wider column count of $B$, which gives a controller or a
learning algorithm more directions in actuator space along which to satisfy a task,
resulting in potentially higher task efficiency at the cost of higher exploration cost
and actuation dimensionality.

\paragraph{Coupling architecture}
Coupling effects depend on how the coupling distributes authority, rather than simply on
its presence or absence. The SDH's J0 coupling at the long fingers slightly worsens the
conditioning ($\kappa(B) = 4.25 > \kappa(J) = 3.84$), since removing the J1 degree of
freedom from independent control reduces the effective rank of $A$. The ACBH's branching
tendon network has the opposite effect on its long fingers, with $\kappa(B) < \kappa(J)$
in all of them. The difference is that the ACBH's network distributes task-space
authority across multiple muscles per joint, whereas the SDH's coupling concentrates two
joints' worth of authority into a single actuator. Networked coupling can therefore
improve the effective conditioning when designed to spread authority across directions,
while sparse coupling tends to concentrate it.

\paragraph{Authority distribution}
This aspect has shown the most consequence in the case of the ACBH's thumb, where
specialized actuator authority produces a $\kappa(B) = 20.03$ outlier from a generally
favourable $\kappa(J) = 4.76$. The main underlying cause for this is the fact that the
human thenar muscle group is optimized for opposition rather than for uniform multi-axis
control of the thumb, which the ACBH's design philosophy replicates faithfully. The SDH
avoids this anisotropy not because its design is biomechanically preferable but because
it has fewer thumb actuators ($m = 5$) acting on fewer DOFs ($n = 5$), each in a
one-to-one mapping.

\paragraph{On the magnitude of the thumb asymmetry}
A distinction should be drawn between the direction of the ACBH thumb's authority asymmetry
and its magnitude. The direction follows from the pattern of
muscle insertion and reproduces a documented anatomical
specialization~\cite{kapandji2007physiology, inouye2014tendon}. The magnitude, on the other hand, is set by the tendon routing and rest-length
parameters encoded in the simulation, so the specific $\kappa(A)\approx30$ and
$\kappa(B)=20.03$ values are sensitive to how those parameters were calibrated. Modifying the tendon lengths would be expected to redistribute authority across the thumb's muscle group and to modulate the reported outliers without changing the qualitative pattern. The values were kept unchanged during this analysis because the emerging patterns were still qualitatively meaningful, and recalibrating the model would have served no additional information with regards to the conclusions drawn.

\subsection{Implications for manipulation tasks}
The four morphological aspects interact in non-trivial ways at the task level. Task
efficiency is approximately balanced when both hands engage their thumbs in a
multi-finger task, since the ACBH's more favourable thumb $\kappa(J)$ is partially offset
by the SDH's superior long-finger $\kappa(J)$, which leaves the system-level $\kappa(J)$
approximately equal ($9.55$ versus $9.45$ for the pinch task).

The morphological strengths of the two hand designs manifest at different scales. The
SDH is structurally homogeneous, in the sense that its $\kappa(B)$ varies from $3.69$ to
$9.45$ across all fingers, and combining fingers in a task does not produce
qualitatively new conditioning behaviour. The ACBH is structurally heterogeneous, since
it contains both the best-conditioned thumb in the study (kinematically) and the
worst-conditioned one (in actuation), and any task that includes the thumb inherits the
latter regardless of the long fingers' favourable contribution.

Whether these structural patterns translate into measurable task performance depends on
the controller's ability to exploit the favourable aspects (long-finger redundancy and
improved $\kappa(B)$ for the ACBH; uniform conditioning across all fingers for the SDH)
while mitigating the unfavourable ones (thumb authority anisotropy for the ACBH; lower
redundancy for the SDH).

\subsection{Implications for control and learning}
The morphological aspects have distinct implications for control design and policy
learning~\cite{pfeifer2006body,hauser2011morphological,gupta2021embodied}. A well-conditioned $J(q)$ implies that
joint-space exploration translates uniformly into task-space coverage, while a
well-conditioned $B(q)$ implies the same for actuator-space exploration, which is the
level at which most practical learning algorithms operate.

Across the four aspects, joint axis geometry favours anatomical morphology only at the
thumb and is irrelevant to learning difficulty at the long fingers. The actuator-to-DOF
ratio sets policy-space dimensionality and, through it, the sample complexity, since the
ACBH's $m = 39$ system actuators present a significantly larger learning problem than the
SDH's $m = 18$. The coupling architecture shows that networked coupling is beneficial for
the long fingers, since the ACBH's branching tendons embed coordination patterns that a
learning algorithm would otherwise have to discover from base principles. Authority
distribution highlights the binding constraint specifically at the ACBH thumb. The high
$\kappa(B) = 20.03$ does not mean the thumb cannot be controlled, since the human hand
demonstrates that anisotropic muscle authority is compatible with extreme dexterity given
sufficient learning capacity, but it means that policy gradient methods will exhibit
ill-conditioned exploration along the MCP and IP flexion directions. This
characteristic requires targeted exploration schedules, action-space regularization, or
curriculum strategies that decouple opposition from MCP and IP control. None of these
adaptations is required for the SDH thumb. Taken together, these patterns show that
morphology acts as a structured set of biases, with different aspects binding at
different scales.

\subsection{Structural predictors of reinforcement learning efficiency}
The structural metrics introduced in this work correspond with specific implications for
the sample efficiency and convergence behaviour of reinforcement learning algorithms,
before training even begins. Three of such connections are highlighted in this
subsection.

\paragraph{Dimensionality and sample complexity}
The actuator count $m$ defines the dimensionality of the action space in which a policy
operates. On-policy methods such as Proximal Policy Optimization (PPO) require gradient
estimates that must be computed across the full action space for each update step, and
increasing this space's dimensionality scales sample complexity approximately linearly at
best (with continuous actions), and exponentially at worst (with a multi-discrete action
space). Off-policy methods such as Deep Deterministic Policy Gradient
(DDPG) and Truncated Quantile Critics (TQC) are more sample-efficient in general, but the
actor's exploration and the critic's approximation are still affected by the size of the
action space~\cite{lillicrap2015ddpg, kuznetsov2020tqc}. The SDH's smaller $m$
($m_{\mathrm{SDH}} = 18$ vs.\ $m_{\mathrm{ACBH}} = 39$) therefore provides a structural
advantage in sample complexity that is independent of their respective kinematic or
conditioning metrics. This advantage is most visible under on-policy algorithms such as
PPO.

\paragraph{Exploration isotropy}
The condition number $\kappa(B)$ of the effective control mapping $B(q) = J(q)A(q)$
(Eq.~\ref{eq:bq}) determines how uniformly the task space is covered during exploration
while using a standard Gaussian exploration noise. A high $\kappa(B)$ implies uneven
exploration, with the task-space directions corresponding to the largest singular values
of $B$ becoming significantly more explored than those corresponding to the smallest ones
on average. For example, the ACBH's $\kappa(B) = 20.03$ at the thumb corresponds to a
roughly $20:1$ ratio between the largest and smallest singular values of the thumb's
actuation-to-task mapping. This results in considerably noisier policy gradient estimates for
certain directions of thumb movement, requiring more samples to achieve equivalent
coverage of the task-space. For tasks that require thumb participation, such as many
object manipulation tasks, this likely results in slower convergence or even a lower
projected success rate than what the thumb's kinematic conditioning ($\kappa(J) = 4.76$)
would suggest. For tasks that primarily require the long fingers, the ACBH's more
favourable $\kappa(B)$ values ($4.10$--$6.57$) may help in offsetting the hand's larger
action space, mainly by embedding coordination patterns that the policy would otherwise
have to discover through long-term trial and error.

\paragraph{Redundancy and stability}
The ACBH's per-finger redundancy ($r = 2$--$3$) results in null-space directions along
which the policy can adjust the fingers' internal posture without modifying the
end-effectors' position and/or velocity. During training, this characteristic can improve
stability by allowing the policy to explore internal configurations that reduce
joint-limit violations or collisions without being penalized by the task's reward
function. The SDH's lower redundancy ($r = 1$--$2$) leaves it with less opportunity to do
so, but having fewer null-space directions has the added effect that fewer motions are
irrelevant to the task, potentially helping to achieve early convergence. Redundancy's
effect on learning stability is ultimately task-dependent, with more straightforward
positioning tasks generally rewarding lower redundancy, and tasks with constraints
associated with contacts/collisions generally rewarding higher redundancy.

\subsection{Interpretation of RL training outcomes}
The training results presented in the Results section are interpreted here against the
structural predictions formulated in the previous subsection.

\paragraph{Reach task (PPO)}
The main structural prediction here was that the SDH's smaller action space
($m_{\mathrm{SDH}} = 18$ vs.\ $m_{\mathrm{ACBH}} = 39$) provides a sample complexity
advantage under PPO, while the ACBH's embedded long-finger coordination
($\kappa(B) < \kappa(J)$ on all four long fingers) may partially offset this. The
direction of the outcome is consistent with the policy-space dimensionality prediction:
the SDH reached a higher success rate in fewer timesteps. The magnitude of the differences
between convergence times and success rates should only be considered as an example,
however, with significantly more repetitions of the two trainings likely providing a more
precise grounds for assessment in this regard. The general behaviour of the two hands is
consistent with the predictions.

\paragraph{BlockRotateZ (DDPG+HER)}
The structural prediction here is that the ACBH thumb's $\kappa(B) = 20.03$ should act as
the dominant conditioning bottleneck, yielding slower convergence and/or a lower
asymptotic success rate than the kinematic conditioning alone would suggest. According to
the results, the two hands perform comparably on this task, with the ACBH demonstrating a
slightly higher success rate, which is not entirely consistent with the prediction. A
plausible explanation is that the Z-axis variant of the block manipulation task mainly
requires the thumb to move and rotate in its well-conditioned axis directions, which can
partially circumvent the existing bottleneck of the thumb being otherwise ill-conditioned.
Another potentially important factor is that for this variant of the block manipulation
task, many target orientations of the cube can be achieved without engaging the thumb too
much, relying instead on wrist joint motions and long-finger manipulation only.

Another observation is that the ACBH's success rate curve is considerably smoother across
the whole training run than the SDH's, which exhibits semi-periodic dips approximately
every $2$ million timesteps. This pattern is consistent with the redundancy prediction:
the ACBH's higher per-finger redundancy ($r = 2$--$3$) may support more stable gradient
estimates during contact-rich phases of training by allowing null-space adjustments that
reduce constraint violations without disrupting task-level progress.

\paragraph{BlockRotateXYZ (DDPG+HER)}
The failure of both hands under DDPG+HER is consistent with the prediction that
high-dimensional action spaces require a more capable algorithm to produce reliable policy
gradient estimates. The result is therefore uninformative with respect to the structural
differences between the two hands specifically.

\paragraph{BlockRotateXYZ (TQC+HER)}
The gap between the two hands under TQC+HER provides the strongest empirical confirmation
of the thumb-bottleneck prediction. A task that requires consistent rotation along all
three axes cannot be reliably decomposed into wrist-mediated or long-finger-mediated
strategies, forcing sustained engagement of the thumb's ill-conditioned actuator-to-task
mapping. The ACBH's substantially lower success rate under these conditions is consistent
with the anisotropic exploration and noisier gradient estimates predicted by
$\kappa(B) = 20.03$ at the thumb.

Taken together, the training outcomes partially confirm the structural predictions derived
from $\kappa(B)$ and $m$. The policy-space dimensionality prediction is supported by the
Reach results under PPO. The dominant-bottleneck prediction for thumb-involving tasks is
confirmed by the BlockRotateXYZ results under TQC+HER, but not by BlockRotateZ,
where task-specific geometry and wrist-mediated compensation appear to offset the
conditioning disadvantage of the ACBH thumb.

\clearpage
\section{Conclusion}
\label{sec:conclusion}
This paper proposed a four-aspect morphological framework, namely joint axis geometry,
actuator-to-DOF ratio, coupling architecture, and authority distribution, and used the
Shadow Dexterous Hand and the Anatomically Correct, Biomechatronic Hand as case studies to
demonstrate how each aspect manifests in standard structural metrics. Parameters for both
hands were derived from their canonical digital representations, with the SDH parameters
obtained from its official URDF and the ACBH parameters obtained from its MuJoCo simulation
model with explicit numerical computation of the moment arm Jacobian.

The contributions of this work can be broken into three components. First, a unified structural framework is presented, in which the kinematics and actuation stages of robotic hands are analysed separately and in composition, through the conditioning of $J$, $A$, and their product $B$, allowing the origin of a coordination bottleneck to be attributed to geometry, actuation and/or the interaction of the two. Second, this framework is applied to two hands representing opposing design philosophies, with all parameters derived directly from their canonical digital representations instead of from idealised models. Third, the structural predictions are tested against reinforcement learning experiments across three different environments and three algorithms, establishing that the framework's predictions are partially validated in training and identifying the conditions in which they are not.

Each aspect produced a distinct outcome. Joint axis geometry yields no kinematic advantage
at the long fingers but can improve thumb conditioning. Actuator-to-DOF ratio provides
per-finger redundancy in proportion to the ratio, but it does not directly improve
conditioning. Coupling architecture favours networked coupling at the long fingers, where
the ACBH achieves $\kappa(B) < \kappa(J)$ on all four. Authority distribution can produce
extreme disparities in joint-level actuator authority, as demonstrated by the $21\times$
imbalance across the ACBH thumb's thenar muscle group, which has a direct consequence for
manipulation tasks. These structural findings translate into specific, testable predictions
for reinforcement-learning efficiency: the SDH's smaller action space ($m = 18$) provides a
sample complexity advantage under on-policy methods, while the ACBH thumb's high $\kappa(B)$
is predicted to produce anisotropic exploration and slower convergence on tasks requiring
consistent thumb participation. Empirical training results from PPO, DDPG+HER, and TQC+HER
experiments are generally consistent with these predictions: the SDH converges faster and
more reliably under straightforward training configurations, while the ACBH could
potentially benefit from implicit and/or explicit reward shaping and curriculum approaches, the
development of which is a clear target for future work.

Taken together, these findings can help in understanding the role of anatomical fidelity in
hand morphology. Rather than providing a uniform structural advantage, anatomical fidelity
invests selectively across the four aspects, while relatively simplified hand designs
sacrifice expressiveness and redundancy in exchange for a uniform actuator authority and a
smaller policy space. Applying the framework detailed in this study to additional hand
designs, including soft-bodied and underactuated grippers, would test its generalizability
beyond the two hands considered here~\cite{odhner2014compliant, grebenstein2012dlr,xu2016biomimetic}.

\clearpage
\section*{CRediT authorship contribution statement}

Zalán Tari: Conceptualization, Methodology, Software, Formal analysis, Investigation, Writing – original draft.\\
Eszter Birtalan:
Resoures, Software,
Writing – review and editing.\\
Péter Polcz: Software, Resources, Writing – review and editing.\\
Miklós Koller: Conceptualization, Supervision, Funding acquisition, Writing – review and editing

\clearpage
\section*{Declaration of competing interest}
The authors declare that they have no known competing financial interests or personal
relationships that could have appeared to influence the work reported in this paper.
\clearpage
\section*{Data availability}
The URDF description of the Shadow Dexterous Hand is publicly available from the Shadow
Robot Company. The ACBH MuJoCo model and the code used to compute the structural metrics
and reinforcement-learning experiments are available from the authors on reasonable request.
\clearpage
\section*{Acknowledgements}
This work was carried out at the Faculty of Information Technology and Bionics, P\'azm\'any
P\'eter Catholic University.\\~\\
Project no. TKP2021-NKTA-66 has been implemented with the support provided by the Ministry of Technology and Industry of Hungary from the National Research, Development and Innovation (NRDI) Fund, financed under the TKP2021-NKTA funding scheme.
OTKA grant no. PD-145902 was supported by the NRDI Fund, and research grant no. BO/00427/25 was supported by the János Bolyai Research Scholarship of the Hungarian Academy of Sciences.\\~\\
Project no. 2023-2.1.2-KDP-2023-00011 has been implemented with the support provided by the Ministry
of Innovation and Technology of Hungary from the National Research, Development and Innovation Fund,
based on the grant award certificate issued by the National Research, Development and Innovation Office.
\clearpage
\appendix

\section{Kinematic parameters of the SDH and ACBH}
\label{app:kin}

\subsection{SDH Denavit--Hartenberg parameters}
The following tables give Craig modified D-H parameters extracted from the SDH URDF model
at zero configuration.

\begin{supptable}[H]
\centering
\begin{tabular}{lcccc}
\toprule
Joint & $a_i$ [m] & $\alpha_i$ [rad] & $d_i$ [m] & $\theta_i$ \\
\midrule
J4 (MCP ab/ad) & 0.0000 & $\frac{\pi}{2}$ & 0.0000 & $\theta_1$ \\
J3 (MCP flex)  & 0.0450 & 0.0000          & 0.0000 & $\theta_2$ \\
J2 (PIP)       & 0.0250 & 0.0000          & 0.0000 & $\theta_3$ \\
\bottomrule
\end{tabular}
\caption{D-H parameters for the index (FF), middle (MF), and ring (RF) fingers of the SDH.}
\end{supptable}

\begin{supptable}[H]
\centering
\begin{tabular}{lcccc}
\toprule
Joint & $a_i$ [m] & $\alpha_i$ [rad] & $d_i$ [m] & $\theta_i$ \\
\midrule
J5 (metacarpal) & 0.0378 & $\frac{\pi}{2}$ & 0.0540 & $\theta_1$ \\
J4 (MCP ab/ad)  & 0.0000 & $\frac{\pi}{2}$ & 0.0000 & $\theta_2$ \\
J3 (MCP flex)   & 0.0450 & 0.0000          & 0.0000 & $\theta_3$ \\
J2 (PIP)        & 0.0250 & 0.0000          & 0.0000 & $\theta_4$ \\
\bottomrule
\end{tabular}
\caption{D-H parameters for the little finger (LF) of the SDH.}
\end{supptable}

\begin{supptable}[H]
\centering
\begin{tabular}{lcccc}
\toprule
Joint & $a_i$ [m] & $\alpha_i$ [rad] & $d_i$ [m] & $\theta_i$ \\
\midrule
J5 (CMC rot)   & 0.0000 & $\frac{\pi}{2}$ & 0.0000 & $\theta_1$ \\
J4 (CMC flex)  & 0.0380 & 0.0000          & 0.0000 & $\theta_2$ \\
J3 (MCP ab/ad) & 0.0000 & $\frac{\pi}{2}$ & 0.0000 & $\theta_3$ \\
J2 (MCP flex)  & 0.0320 & 0.0000          & 0.0000 & $\theta_4$ \\
J1 (IP)        & 0.0226 & 0.0000          & 0.0000 & $\theta_5$ \\
\bottomrule
\end{tabular}
\caption{D-H parameters for the thumb (TH) of the SDH.}
\end{supptable}

\subsection{SDH PoE screw axes}

\begin{supptable}[H]
\centering\small
\begin{tabular}{lrrrrrr}
\toprule
Joint & $\omega_1$ & $\omega_2$ & $\omega_3$ & $v_1$\,[m] & $v_2$\,[m] & $v_3$\,[m] \\
\midrule
\multicolumn{7}{l}{\textit{Index / Middle / Ring fingers}} \\
J4 (MCP ab/ad) &  0.000 & $-$1.000 &  0.000 & 0.0950 & 0.0000 & $-$0.0330 \\
J3 (MCP flex)  &  1.000 &  0.000 &  0.000  & 0.0000 & 0.0950 &  0.0000 \\
J2 (PIP)       &  1.000 &  0.000 &  0.000  & 0.0000 & 0.1400 &  0.0000 \\
J1 (DIP)       &  1.000 &  0.000 &  0.000  & 0.0000 & 0.1650 &  0.0000 \\
\midrule
\multicolumn{7}{l}{\textit{Little finger}} \\
J5 (metacarpal) & 0.5736 &  0.000 & 0.8191 & 0.0000 & 0.0389 &  0.0000 \\
J4 (MCP ab/ad)  & 0.000 &  1.000 & 0.000  & $-$0.0866 & 0.0000 & $-$0.0330 \\
J3 (MCP flex)   & 1.000 &  0.000 & 0.000  & 0.0000 & 0.0866 &  0.0000 \\
J2 (PIP)        & 1.000 &  0.000 & 0.000  & 0.0000 & 0.1316 &  0.0000 \\
J1 (DIP)        & 1.000 &  0.000 & 0.000  & 0.0000 & 0.1566 &  0.0000 \\
\midrule
\multicolumn{7}{l}{\textit{Thumb}} \\
J5 (CMC rot)   & $-$0.7071 &  0.000 & $-$0.7071 & 0.0061 & 0.0035 & $-$0.0061 \\
J4 (CMC flex)  &  0.7071 &  0.000 & $-$0.7071 & 0.0061 & 0.0445 &  0.0061 \\
J3 (MCP ab/ad) &  0.7071 &  0.000 & $-$0.7071 & 0.0061 & 0.0825 &  0.0061 \\
J2 (MCP flex)  &  0.000 & $-$1.000 &  0.000   & 0.0559 & 0.0000 & $-$0.0609 \\
J1 (IP)        &  0.000 & $-$1.000 &  0.000   & 0.0785 & 0.0000 & $-$0.0835 \\
\bottomrule
\end{tabular}
\caption{SDH PoE screw axes
$\hat{\xi}_i = [\omega_i;\, p_i \times \omega_i]$ in the palm-fixed space frame at zero
configuration, derived from the official URDF.}
\label{tab:poe_sdh}
\end{supptable}

\subsection{ACBH Denavit--Hartenberg parameters}
The following tables give Craig modified D-H parameters extracted from the ACBH MuJoCo model
at zero configuration. The distal joint of each finger (DIP for long fingers, IP for the
thumb) is absorbed into the end-effector transform $T_0$ and is not listed separately. Small
residual $\alpha$ values can be attributed to the oblique joint axes in the anatomical model,
and are not numerical inconsistencies.

\begin{supptable}[H]
\centering
\begin{tabular}{lcccc}
\toprule
Joint & $a_i$\,[m] & $\alpha_i$\,[rad] & $d_i$\,[m] & $\theta_i$ \\
\midrule
MCP\_Yaw   & 0.0000 & $\frac{\pi}{2}$ & 0.0000 & $\theta_1$ \\
MCP\_Roll  & 0.0000 & $\frac{\pi}{2}$ & 0.0000 & $\theta_2$ \\
MCP\_Pitch & 0.0500 & 0.0174 & 0.0000 & $\theta_3$ \\
PIP\_Pitch & 0.0300 & 0.0000 & 0.0000 & $\theta_4$ \\
\bottomrule
\end{tabular}
\caption{D-H parameters for the index finger (F2) of the ACBH.}
\label{tab:dh_acbh_f2}
\end{supptable}

\begin{supptable}[H]
\centering
\begin{tabular}{lcccc}
\toprule
Joint & $a_i$\,[m] & $\alpha_i$\,[rad] & $d_i$\,[m] & $\theta_i$ \\
\midrule
MCP\_Yaw   & 0.0000 & $\frac{\pi}{2}$ & 0.0000 & $\theta_1$ \\
MCP\_Roll  & 0.0000 & $\frac{\pi}{2}$ & 0.0000 & $\theta_2$ \\
MCP\_Pitch & 0.0520 & 0.0174 & 0.0000 & $\theta_3$ \\
PIP\_Pitch & 0.0310 & 0.0000 & 0.0000 & $\theta_4$ \\
\bottomrule
\end{tabular}
\caption{D-H parameters for the middle finger (F3) of the ACBH.}
\label{tab:dh_acbh_f3}
\end{supptable}

\begin{supptable}[H]
\centering
\begin{tabular}{lcccc}
\toprule
Joint & $a_i$\,[m] & $\alpha_i$\,[rad] & $d_i$\,[m] & $\theta_i$ \\
\midrule
MCP\_Yaw   & 0.0000 & $\frac{\pi}{2}$ & 0.0000 & $\theta_1$ \\
MCP\_Roll  & 0.0000 & $\frac{\pi}{2}$ & 0.0000 & $\theta_2$ \\
MCP\_Pitch & 0.0510 & 0.0349 & 0.0000 & $\theta_3$ \\
PIP\_Pitch & 0.0290 & 0.0000 & 0.0000 & $\theta_4$ \\
\bottomrule
\end{tabular}
\caption{D-H parameters for the ring finger (F4) of the ACBH.}
\label{tab:dh_acbh_f4}
\end{supptable}

\begin{supptable}[H]
\centering
\begin{tabular}{lcccc}
\toprule
Joint & $a_i$\,[m] & $\alpha_i$\,[rad] & $d_i$\,[m] & $\theta_i$ \\
\midrule
MCP\_Yaw   & 0.0000 & $\frac{\pi}{2}$ & 0.0000 & $\theta_1$ \\
MCP\_Roll  & 0.0000 & $\frac{\pi}{2}$ & 0.0000 & $\theta_2$ \\
MCP\_Pitch & 0.0460 & 0.0174 & 0.0000 & $\theta_3$ \\
PIP\_Pitch & 0.0260 & 0.0000 & 0.0000 & $\theta_4$ \\
\bottomrule
\end{tabular}
\caption{D-H parameters for the little finger (F5) of the ACBH.}
\label{tab:dh_acbh_f5}
\end{supptable}

\begin{supptable}[H]
\centering
\begin{tabular}{lcccc}
\toprule
Joint & $a_i$\,[m] & $\alpha_i$\,[rad] & $d_i$\,[m] & $\theta_i$ \\
\midrule
TCMC\_Yaw   & 0.0000 & $\frac{\pi}{2}$ & 0.0000 & $\theta_1$ \\
TCMC\_Pitch & 0.0500 & 1.5202 & 0.0000 & $\theta_2$ \\
MCP\_Yaw    & 0.0000 & $\frac{\pi}{2}$ & 0.0000 & $\theta_3$ \\
MCP\_Roll   & 0.0000 & $\frac{\pi}{2}$ & 0.0000 & $\theta_4$ \\
MCP\_Pitch  & 0.0400 & 0.0000 & 0.0000 & $\theta_5$ \\
\bottomrule
\end{tabular}
\caption{D-H parameters for the thumb (T1) of the ACBH.}
\label{tab:dh_acbh_t1}
\end{supptable}

\subsection{ACBH PoE screw axes}

\begin{supptable}[H]
\centering\small
\begin{tabular}{lrrrrrr}
\toprule
Joint & $\omega_1$ & $\omega_2$ & $\omega_3$ & $v_1$\,[m] & $v_2$\,[m] & $v_3$\,[m] \\
\midrule
MCP\_Yaw   &  0.035 &  0.914 &  0.405 & $-$0.074 & $-$0.007 &  0.022 \\
MCP\_Roll  & $-$0.015 & $-$0.405 &  0.914 &  0.030 & $-$0.023 & $-$0.010 \\
MCP\_Pitch &  0.999 & $-$0.038 & $-$0.001 &  0.003 &  0.080 &  0.002 \\
PIP\_Pitch &  0.999 & $-$0.054 & $-$0.008 &  0.007 &  0.125 &  0.022 \\
DIP\_Pitch &  0.999 & $-$0.054 & $-$0.008 &  0.008 &  0.145 &  0.044 \\
\bottomrule
\end{tabular}
\caption{ACBH PoE screw axes for the index finger (F2).}
\label{tab:poe_acbh_index}
\end{supptable}

\begin{supptable}[H]
\centering\small
\begin{tabular}{lrrrrrr}
\toprule
Joint & $\omega_1$ & $\omega_2$ & $\omega_3$ & $v_1$\,[m] & $v_2$\,[m] & $v_3$\,[m] \\
\midrule
MCP\_Yaw   &  0.000 &  0.866 &  0.500 & $-$0.071 &  0.000 &  0.000 \\
MCP\_Roll  &  0.000 & $-$0.500 &  0.866 &  0.041 &  0.000 &  0.000 \\
MCP\_Pitch &  1.000 &  0.000 &  0.000 &  0.000 &  0.082 &  0.000 \\
PIP\_Pitch &  1.000 &  0.015 &  0.009 & $-$0.002 &  0.127 &  0.026 \\
DIP\_Pitch &  1.000 &  0.015 &  0.009 & $-$0.003 &  0.143 &  0.053 \\
\bottomrule
\end{tabular}
\caption{ACBH PoE screw axes for the middle finger (F3).}
\label{tab:poe_acbh_middle}
\end{supptable}

\begin{supptable}[H]
\centering\small
\begin{tabular}{lrrrrrr}
\toprule
Joint & $\omega_1$ & $\omega_2$ & $\omega_3$ & $v_1$\,[m] & $v_2$\,[m] & $v_3$\,[m] \\
\midrule
MCP\_Yaw   & $-$0.056 &  0.905 &  0.422 & $-$0.069 &  0.005 & $-$0.019 \\
MCP\_Roll  &  0.045 & $-$0.419 &  0.907 &  0.032 &  0.023 &  0.009 \\
MCP\_Pitch &  0.997 &  0.070 & $-$0.018 & $-$0.005 &  0.075 & $-$0.001 \\
PIP\_Pitch &  0.995 &  0.101 & $-$0.003 & $-$0.012 &  0.121 &  0.020 \\
DIP\_Pitch &  0.995 &  0.101 & $-$0.003 & $-$0.014 &  0.137 &  0.044 \\
\bottomrule
\end{tabular}
\caption{ACBH PoE screw axes for the ring finger (F4).}
\label{tab:poe_acbh_ring}
\end{supptable}

\begin{supptable}[H]
\centering\small
\begin{tabular}{lrrrrrr}
\toprule
Joint & $\omega_1$ & $\omega_2$ & $\omega_3$ & $v_1$\,[m] & $v_2$\,[m] & $v_3$\,[m] \\
\midrule
MCP\_Yaw   & $-$0.144 &  0.882 &  0.448 & $-$0.061 &  0.010 & $-$0.039 \\
MCP\_Roll  &  0.033 & $-$0.448 &  0.893 &  0.027 &  0.041 &  0.020 \\
MCP\_Pitch &  0.989 &  0.144 &  0.036 & $-$0.010 &  0.068 & $-$0.003 \\
PIP\_Pitch &  0.986 &  0.159 &  0.044 & $-$0.018 &  0.108 &  0.017 \\
DIP\_Pitch &  0.986 &  0.159 &  0.044 & $-$0.021 &  0.121 &  0.040 \\
\bottomrule
\end{tabular}
\caption{ACBH PoE screw axes for the little finger (F5).}
\label{tab:poe_acbh_little}
\end{supptable}

\begin{supptable}[H]
\centering\small
\begin{tabular}{lrrrrrr}
\toprule
Joint & $\omega_1$ & $\omega_2$ & $\omega_3$ & $v_1$\,[m] & $v_2$\,[m] & $v_3$\,[m] \\
\midrule
TCMC\_Yaw   &  0.850 &  0.396 & $-$0.346 &  0.004 &  0.007 &  0.019 \\
TCMC\_Pitch &  0.142 & $-$0.807 & $-$0.574 &  0.000 &  0.018 & $-$0.025 \\
MCP\_Yaw    &  0.991 &  0.132 & $-$0.029 & $-$0.003 &  0.034 &  0.036 \\
MCP\_Roll   &  0.098 & $-$0.554 &  0.827 & $-$0.006 & $-$0.044 & $-$0.029 \\
MCP\_Pitch  &  0.094 & $-$0.822 & $-$0.562 &  0.043 &  0.035 & $-$0.044 \\
IP\_Pitch   &  0.094 & $-$0.822 & $-$0.562 &  0.082 &  0.041 & $-$0.046 \\
\bottomrule
\end{tabular}
\caption{ACBH PoE screw axes for the thumb (T1).}
\label{tab:poe_acbh_thumb}
\end{supptable}

\section{Reinforcement learning training details}
\label{app:rl}

\subsection{Success-rate curves}

\begin{suppfigure}[H]
    \centering
    \includegraphics[width=0.9\linewidth]{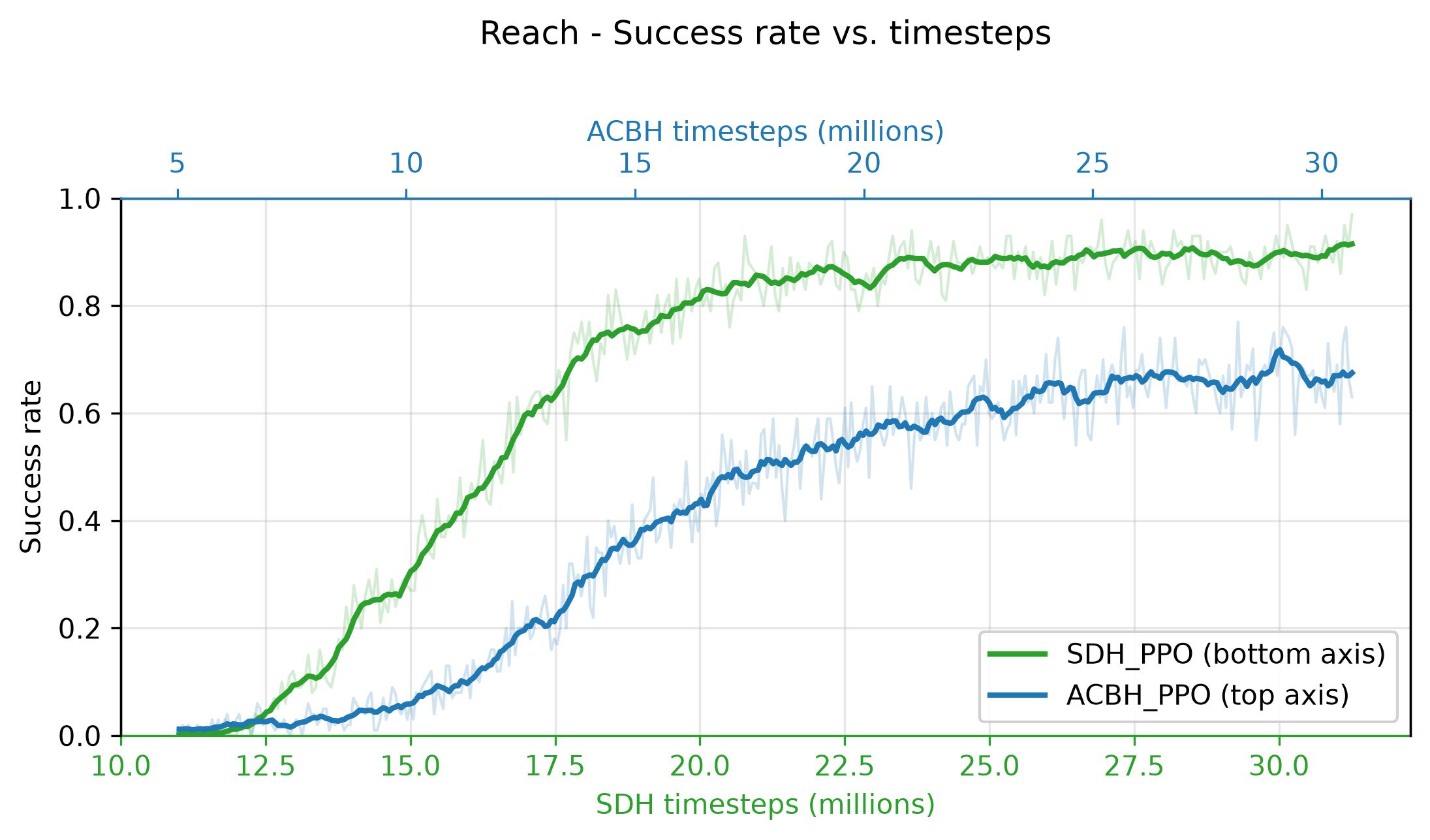}
    \caption{Success rates of the SDH and ACBH over time in the Reach task environment,
    trained with PPO. The smoothed values are computed using an exponential moving average
    with weight $\alpha=0.6$.}
    \label{fig:reach_success_rate}
\end{suppfigure}

\begin{suppfigure}[H]
    \centering
    \includegraphics[width=0.9\linewidth]{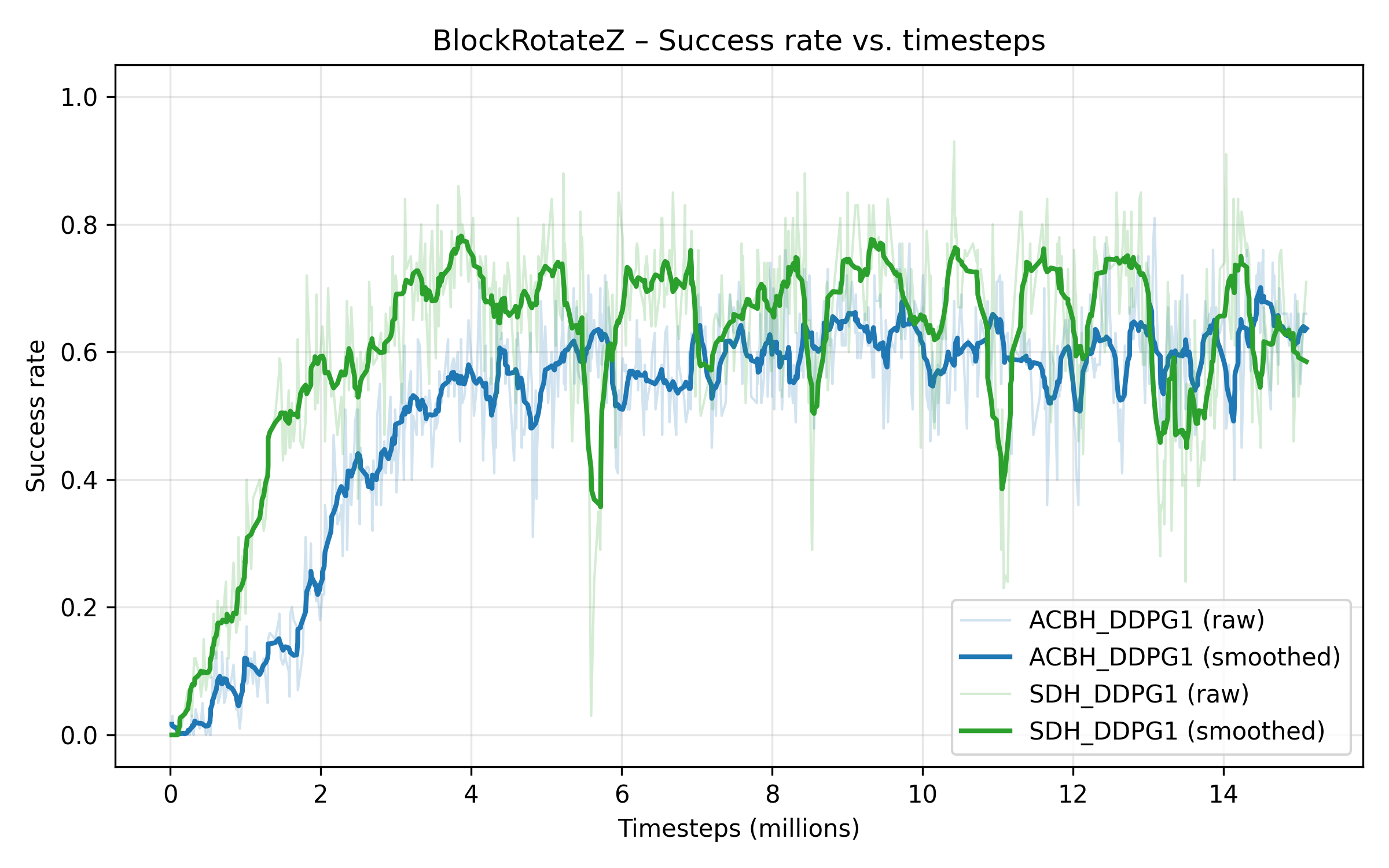}
    \caption{Success rates of the SDH and ACBH over time in the BlockRotateZ task
    environment, trained with DDPG+HER. The smoothed values are computed using an
    exponential moving average with weight $\alpha=0.6$.}
    \label{fig:rotate_z_success_rate}
\end{suppfigure}

\begin{suppfigure}[H]
    \centering
    \includegraphics[width=0.9\linewidth]{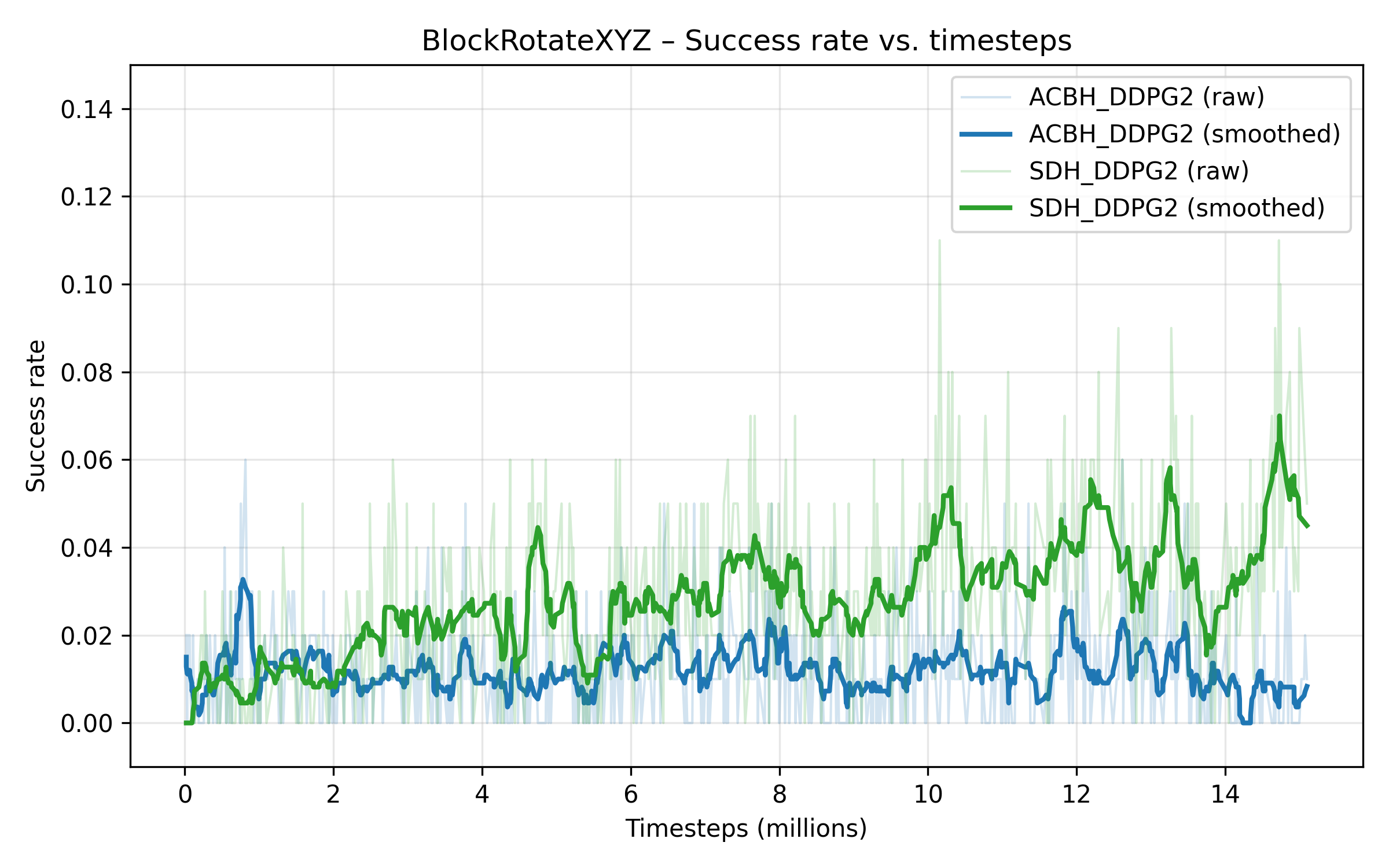}
    \caption{Success rates of the SDH and ACBH over time in the BlockRotateXYZ task
    environment, trained with DDPG+HER. The smoothed values are computed using an
    exponential moving average with weight $\alpha=0.6$.}
    \label{fig:rotate_xyz_ddpg_success_rate}
\end{suppfigure}

\begin{suppfigure}[H]
    \centering
    \includegraphics[width=0.9\linewidth]{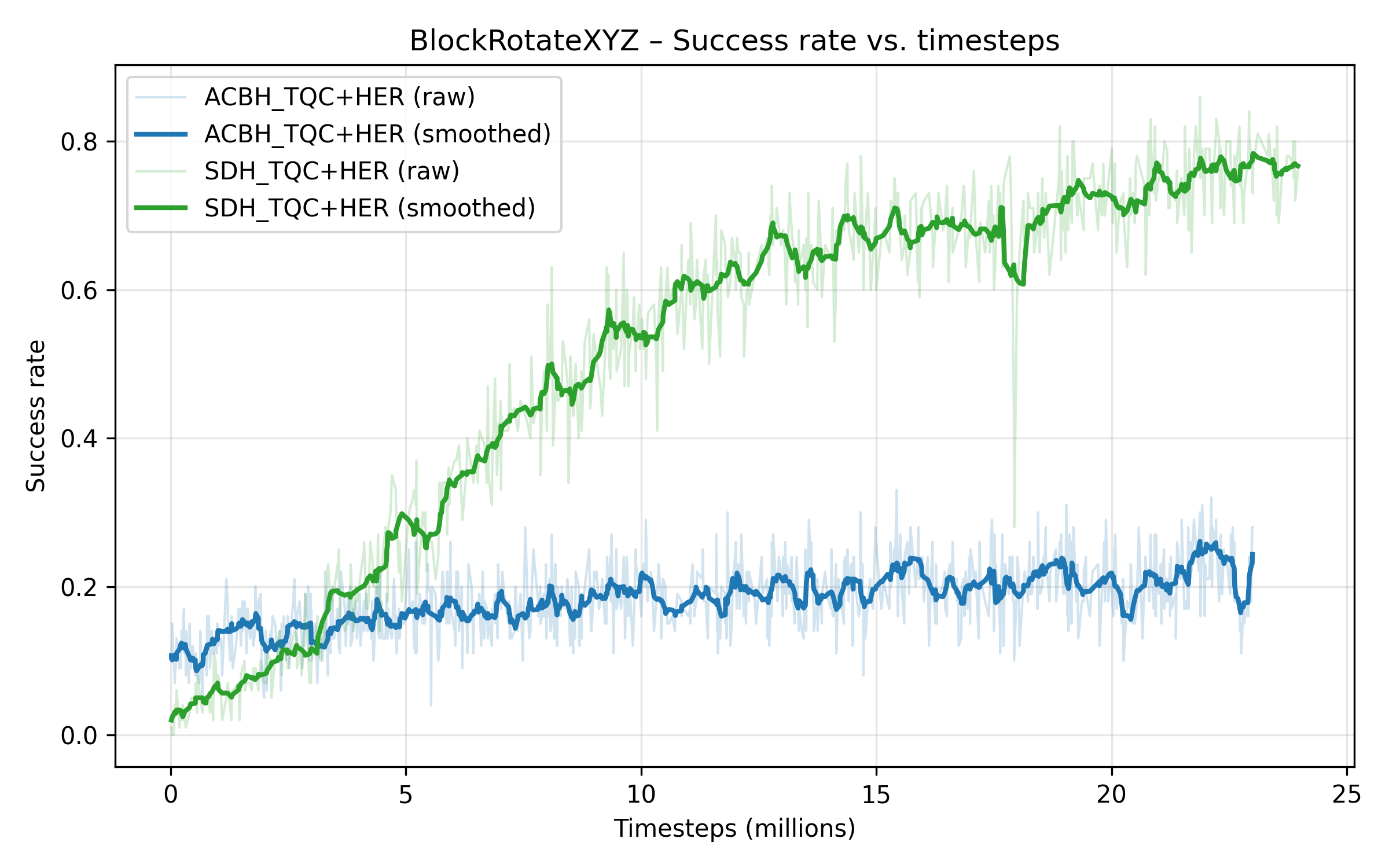}
    \caption{Success rates of the SDH and ACBH over time in the BlockRotateXYZ task
    environment (with rotation\_threshold set to 0.4 radians), trained with TQC+HER. The
    smoothed values are computed using an exponential moving average with weight
    $\alpha=0.6$.}
    \label{fig:rotate_xyz_tqc_success_rate}
\end{suppfigure}

\subsection{Computational environment}

The training runs mentioned during this work were carried out locally, using a workstation equipped with an NVIDIA GeForce
RTX 5060 Ti (16\,GB), an Intel Core i5-14400F processor, and 32\,GB of DDR4
memory. All experiments used Stable-Baselines3 2.7.0~\cite{raffin2021sb3}
with the MuJoCo 3.10.0 physics engine~\cite{todorov2012mujoco}. Environments
were vectorised across 16 parallel instances for the Reach task and 19 for both
block manipulation variants.

\subsection{Hyperparameters}

\noindent\textbf{PPO hyperparameters:}
\begin{itemize}
    \item Policy: \texttt{MultiInputPolicy}
    \item learning\_rate = cosine\_schedule(3e-4, 6e-5)
    \item n\_steps = 4096
    \item batch\_size = 512
    \item n\_epochs = 10
    \item gamma = 0.98
    \item gae\_lambda = 0.95
    \item clip\_range = 0.2
    \item ent\_coef = 0.005
    \item max\_grad\_norm = 0.7
    \item normalize\_advantage = True
\end{itemize}
\textbf{Other parameters affecting the environment or normalization:}
\begin{itemize}
    \item num\_envs = 16
    \item max\_episode\_steps = 100
    \item VecNormalize/norm\_obs = True
    \item VecNormalize/norm\_reward = True
    \item VecNormalize/clip\_obs = 10.0
\end{itemize}

\noindent\textbf{DDPG+HER hyperparameters:}
\begin{itemize}
    \item Policy: \texttt{MultiInputPolicy}
    \item replay\_buffer\_class = HerReplayBuffer
    \item replay\_buffer\_kwargs = \{ n\_sampled\_goal = 4, goal\_selection\_strategy = "future" \}
    \item learning\_rate = 5e-4
    \item gamma = 0.996 ($=1-\frac{1}{250\,(\text{max\_episode\_steps})}$)
    \item tau = 0.05
    \item buffer\_size = 1e6
    \item batch\_size = 256
    \item gradient\_steps = 4
    \item train\_freq = (4, "step")
    \item learning\_starts = 250 * 19 * 10 ($=$ max\_episode\_steps * num\_envs * 10)
    \item policy\_kwargs = \{ net\_arch = [256, 256, 256], n\_critics = 2 \}
\end{itemize}
\textbf{Other parameters affecting the environment or normalization:}
\begin{itemize}
    \item num\_envs = 19
    \item max\_episode\_steps = 250
    \item VecNormalize/norm\_obs = True
    \item VecNormalize/norm\_reward = False
    \item VecNormalize/clip\_obs = 5.0
\end{itemize}

\noindent\textbf{TQC+HER hyperparameters:}
\begin{itemize}
    \item Policy: \texttt{MultiInputPolicy}
    \item replay\_buffer\_class = HerReplayBuffer
    \item replay\_buffer\_kwargs = \{ n\_sampled\_goal = 4, goal\_selection\_strategy = "future" \}
    \item learning\_rate = 3e-4
    \item gamma = 0.99 ($=1-\frac{1}{100\,(\text{max\_episode\_steps})}$)
    \item tau = 0.05
    \item buffer\_size = 1e6
    \item batch\_size = 512
    \item gradient\_steps = 4
    \item train\_freq = 2
    \item learning\_starts = 100 * 19 * 10 ($=$ max\_episode\_steps * num\_envs * 10)
    \item top\_quantiles\_to\_drop\_per\_net = 2
    \item policy\_kwargs = \{ net\_arch = [256, 256], n\_quantiles = 25 \}
\end{itemize}
\textbf{Other parameters affecting the environment or normalization:}
\begin{itemize}
    \item num\_envs = 19
    \item max\_episode\_steps = 100
    \item VecNormalize/norm\_obs = True
    \item VecNormalize/norm\_reward = False
    \item VecNormalize/clip\_obs = 5.0
\end{itemize}

\clearpage

\bibliographystyle{cas-model1-num-names}
\bibliography{references}

\end{document}